\documentclass[10pt,journal,compsoc]{IEEEtran}
\newif\ifpeerreview
\newif\ifsupplement
\newif\ifmain  % everything other than the supplement

\peerreviewfalse
\maintrue
\supplementtrue

\usepackage[nocompress]{cite}
\usepackage{url}
\usepackage{amsmath,amssymb,graphicx}
\usepackage{multirow,xcolor}
\usepackage[export]{adjustbox}
\usepackage{booktabs}

\usepackage[switch]{lineno}
\usepackage[hidelinks]{hyperref}
\usepackage{cleveref}
\usepackage{bbm}
\usepackage{tikz}

\newcommand{\paperID}{17}

\title{ThermalNeRF: Thermal Radiance Fields}

\author{Yvette~Y.~Lin*, %~\IEEEmembership{Member,~IEEE,}
        Xin-Yi~Pan*,
        Sara~Fridovich-Keil,
        and~Gordon~Wetzstein%~\IEEEmembership{Life~Fellow,~IEEE}% <-this % stops a space
\IEEEcompsocitemizethanks{\IEEEcompsocthanksitem Y. Y. Lin is with the Department
of Computer Science, Stanford University, Stanford, CA.
\IEEEcompsocthanksitem X. Pan, S. Fridovich-Keil, G. Wetzstein are with Department of Electrical Engineering, Stanford University, Stanford, CA.
\IEEEcompsocthanksitem E-mail: \{yvelin, xinyipan, sarafk, gordonwz\}@stanford.edu
}% <-this % stops an unwanted space
}

\newcommand{\R}{\mathbb{R}}

\newcommand{\X}{\mathcal{X}}
\newcommand{\Loss}{\mathcal{L}}
\newcommand{\rgb}{\text{rgb}}
\newcommand{\therm}{\text{therm}}
\newcommand{\crosschannel}{\text{cc}}
\newcommand{\density}{\text{$\sigma$}}
\newcommand{\tv}{\text{tv}}
\newcommand{\gt}{\text{gt}}
\renewcommand{\vec}[1]{\mathbf{#1}}

\newcommand{\plotfourtrimcenter}[1]{%
  \adjincludegraphics[trim={{0.2\width} {0.2\height} {0.2\width} {0.2\height}}, clip, width=0.23\linewidth]{#1}%
}

\newcommand{\plotscenecropped}[3]{
\begin{figure*}[h]
\centering
\begin{tabular}{l@{~~}c@{}c@{}c@{}c@{}}
\rotatebox{90}{~~~~~~~~~~~~RGB} & \plotfourtrimcenter{figs/#1/ground_truth_rgb/frame_#2.jpg} & 
\plotfourtrimcenter{figs/#1/ours/frame_#2_rgb.jpg} &
\plotfourtrimcenter{figs/#1/combined/frame_#2_rgb.jpg} &
\plotfourtrimcenter{figs/#1/separate/frame_#2_rgb.jpg} \\
\rotatebox{90}{~~~~~~~~~~Thermal} & \plotfourtrimcenter{figs/#1/ground_truth_thermal/frame_#2.jpg} & 
\plotfourtrimcenter{figs/#1/ours/frame_#2_thermal.jpg} &
\plotfourtrimcenter{figs/#1/combined/frame_#2_thermal.jpg} &
\plotfourtrimcenter{figs/#1/separate/frame_#2_thermal.jpg} \\
& Ground Truth & Ours & Nerfacto \{RGBT\} & Nerfacto \{RGB\}\{T\}
\end{tabular}
\caption{#3}
\label{fig:#1}
\end{figure*}
}

\newcommand{\plotfour}[1]{%
  \adjincludegraphics[trim={{0\width} {0\height} {0\width} {0\height}}, clip, width=0.23\linewidth]{#1}%
}

\newcommand{\plotscene}[3]{
\begin{figure*}[!h]
\centering
\begin{tabular}{l@{~~}c@{}c@{}c@{}c@{}}
\rotatebox{90}{~~~~~~~~~~~~RGB} & \plotfour{figs/#1/ground_truth_rgb/frame_#2.jpg} & 
\plotfour{figs/#1/ours/frame_#2_rgb.jpg} &
\plotfour{figs/#1/combined/frame_#2_rgb.jpg} &
\plotfour{figs/#1/separate/frame_#2_rgb.jpg} \\
\rotatebox{90}{~~~~~~~~~~Thermal} & \plotfour{figs/#1/ground_truth_thermal/frame_#2.jpg} & 
\plotfour{figs/#1/ours/frame_#2_thermal.jpg} &
\plotfour{figs/#1/combined/frame_#2_thermal.jpg} &
\plotfour{figs/#1/separate/frame_#2_thermal.jpg} \\
& Ground Truth & Ours & Nerfacto \{RGBT\} & Nerfacto \{RGB\}\{T\}
\end{tabular}
\caption{#3}
\label{fig:#1}
\end{figure*}
}

\newcommand{\drawbox}[5]{\begin{tikzpicture}
    \node[anchor=south west,inner sep=0] (image) at (0,0) {#1};
    \begin{scope}[x={(image.south east)},y={(image.north west)}]
        \draw[red,thick,rounded corners] (#2,#3) rectangle (#4,#5);
    \end{scope}
\end{tikzpicture}
}

\begin{document}

\ifmain

\IEEEtitleabstractindextext{%
\begin{abstract}
Thermal imaging has a variety of applications, from agricultural monitoring to building inspection to imaging under poor visibility, such as in low light, fog, and rain.
However, reconstructing thermal scenes in 3D presents several challenges due to the comparatively lower resolution and limited features present in long-wave infrared (LWIR) images. To overcome these challenges, we propose a unified framework for scene reconstruction from a set of LWIR and RGB images, using a multispectral radiance field to represent a scene viewed by both visible and infrared cameras, thus leveraging information across both spectra. We calibrate the RGB and infrared cameras with respect to each other, as a preprocessing step using a simple calibration target. We demonstrate our method on real-world sets of RGB and LWIR photographs captured from a handheld thermal camera, showing the effectiveness of our method at scene representation across the visible and infrared spectra.
We show that our method is capable of thermal super-resolution, as well as visually removing obstacles to reveal objects that are occluded in either the RGB or thermal channels.
Please see \url{https://yvette256.github.io/thermalnerf/} for video results as well as our code and dataset release.
\end{abstract}

}
\fi

% Make Title
\ifpeerreview
\linenumbers \linenumbersep 15pt\relax
\author{Paper ID \paperID\IEEEcompsocitemizethanks{\IEEEcompsocthanksitem This paper is under review for ICCP 2024 and the PAMI special issue on computational photography. Do not distribute.}}
\markboth{Anonymous ICCP 2024 submission ID \paperID}%
{}
\fi

\ifmain
\maketitle
\thispagestyle{empty}

% The first section title should be wrapped inside a \IEEEraisesectionheading as follows.
\IEEEraisesectionheading{
  \section{Introduction}  
}
\label{sec:intro}

Thermal imaging exposes features of our world that are invisible to the naked eye, and to RGB cameras recording visible light. By capturing long-wave infrared (LWIR) light, in the 8--14 $\mu$m wavelength range, thermal cameras expose heat sources and material properties, and can see in the dark as well as through many occlusive media such as smoke. These properties make thermal imaging a valuable tool in a wide range of applications including water and air pollution monitoring \cite{fuentes2021urban, lega2010aerial, iwasaki2023real, pyykonen2016multi}, search and rescue \cite{rudol2008human, rodin2018object}, burn severity triage \cite{goel2020prospective, jaspers2017flir}, surveillance and defense \cite{torresan2004advanced, akula2011thermal, wong2009effective}, agriculture and vegetation monitoring \cite{jurado2022remote, nasi2018remote, miyoshi2020novel, lee2018analyzing, fuentes2019non, carrasco2020performance, hernandez2019early}, and infrastructure inspection \cite{concrete_testing, jackson1998leak}.

In many of these applications, multiple thermal images are collected from different viewpoints during the course of inspection or exploration. These applications stand to benefit from access to 3D thermal field reconstructions that combine these multi-view thermal images into a unified and consistent 3D thermal volume; this task of 3D thermal field recovery is the focus of our work. For example, in \cref{fig:skydio} we show example renderings from our reconstruction of a large crane structure based on RGB and thermal images collected autonomously by a Skydio drone. Combining drone-based thermal and visible imaging with our multi-spectral 3D reconstruction may help automate and accelerate otherwise difficult, tedious, or dangerous tasks including this example of infrastructure inspection.

\begin{figure}[h]
    \centering
    \includegraphics[height=0.23\linewidth]{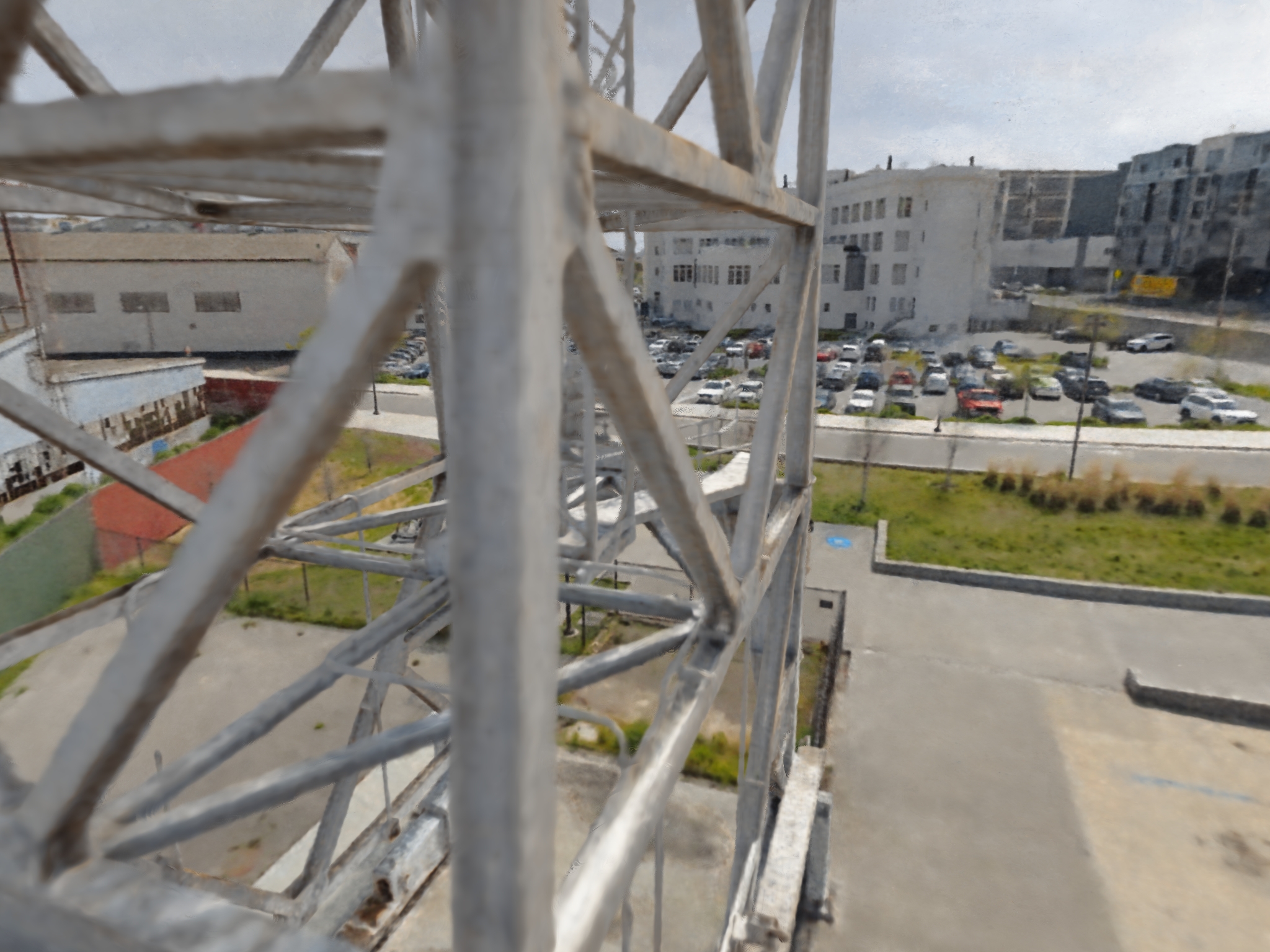}
    \includegraphics[height=0.23\linewidth]{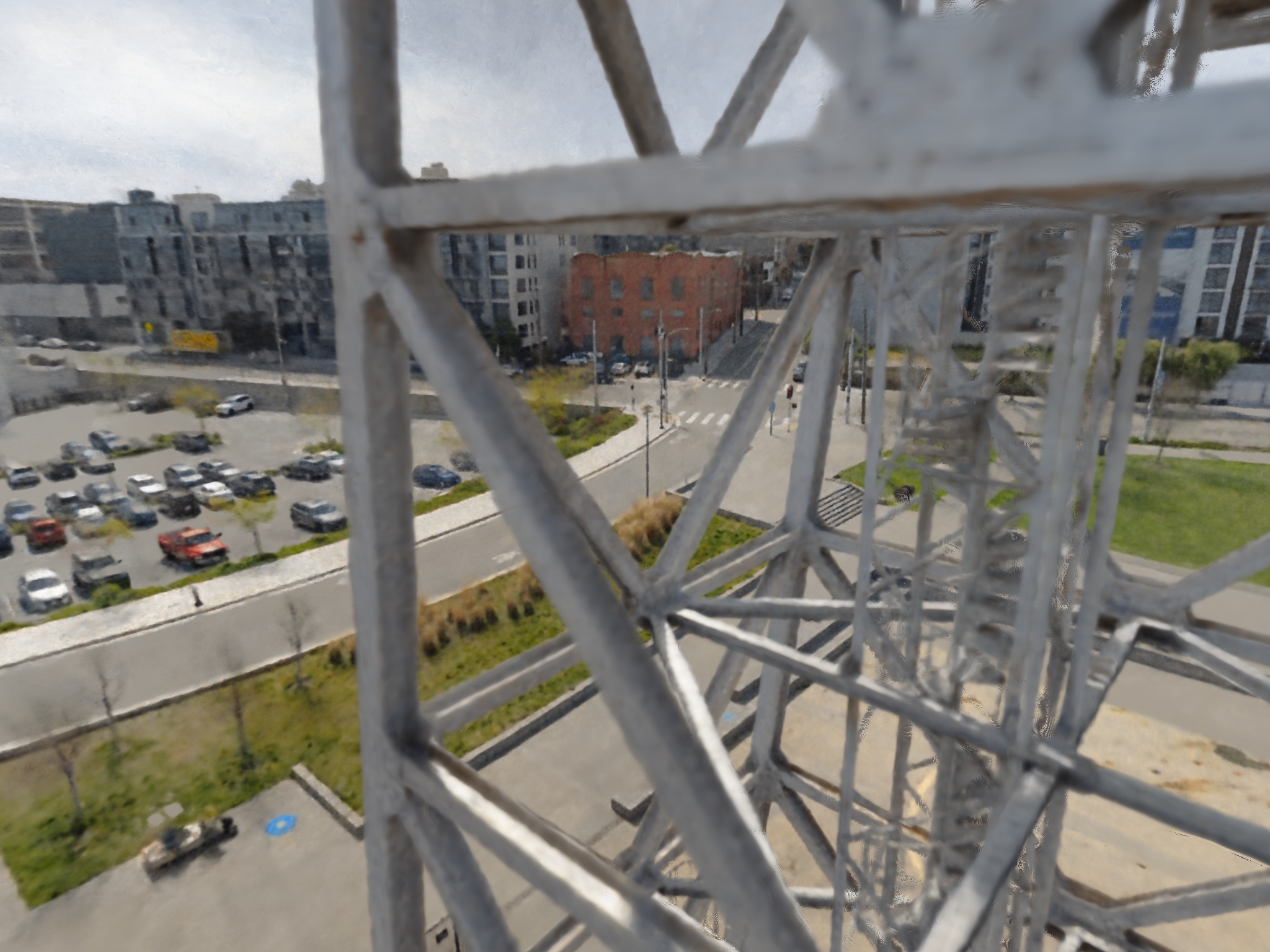}
    \includegraphics[height=0.23\linewidth]{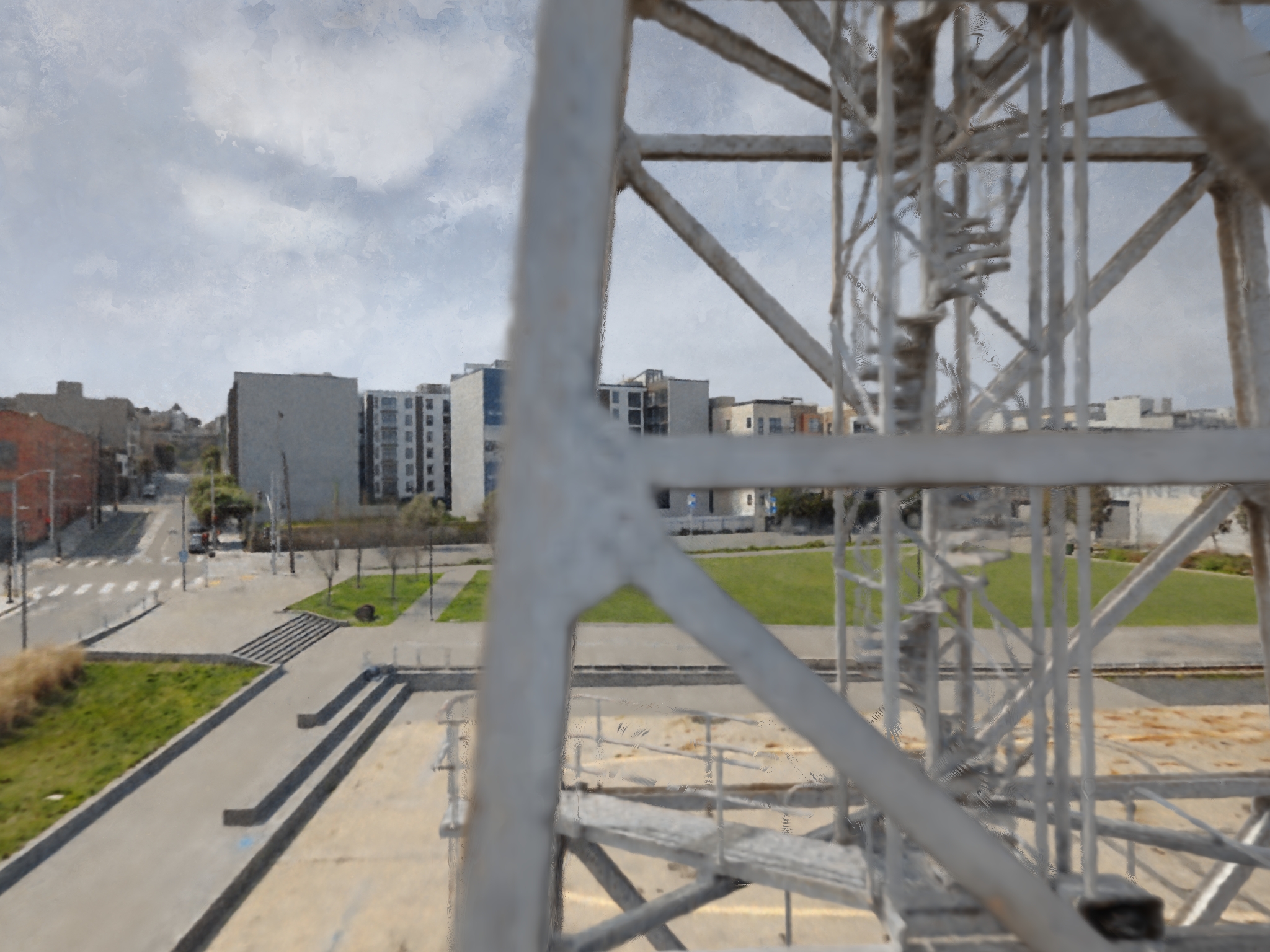}
    \includegraphics[height=0.23\linewidth]{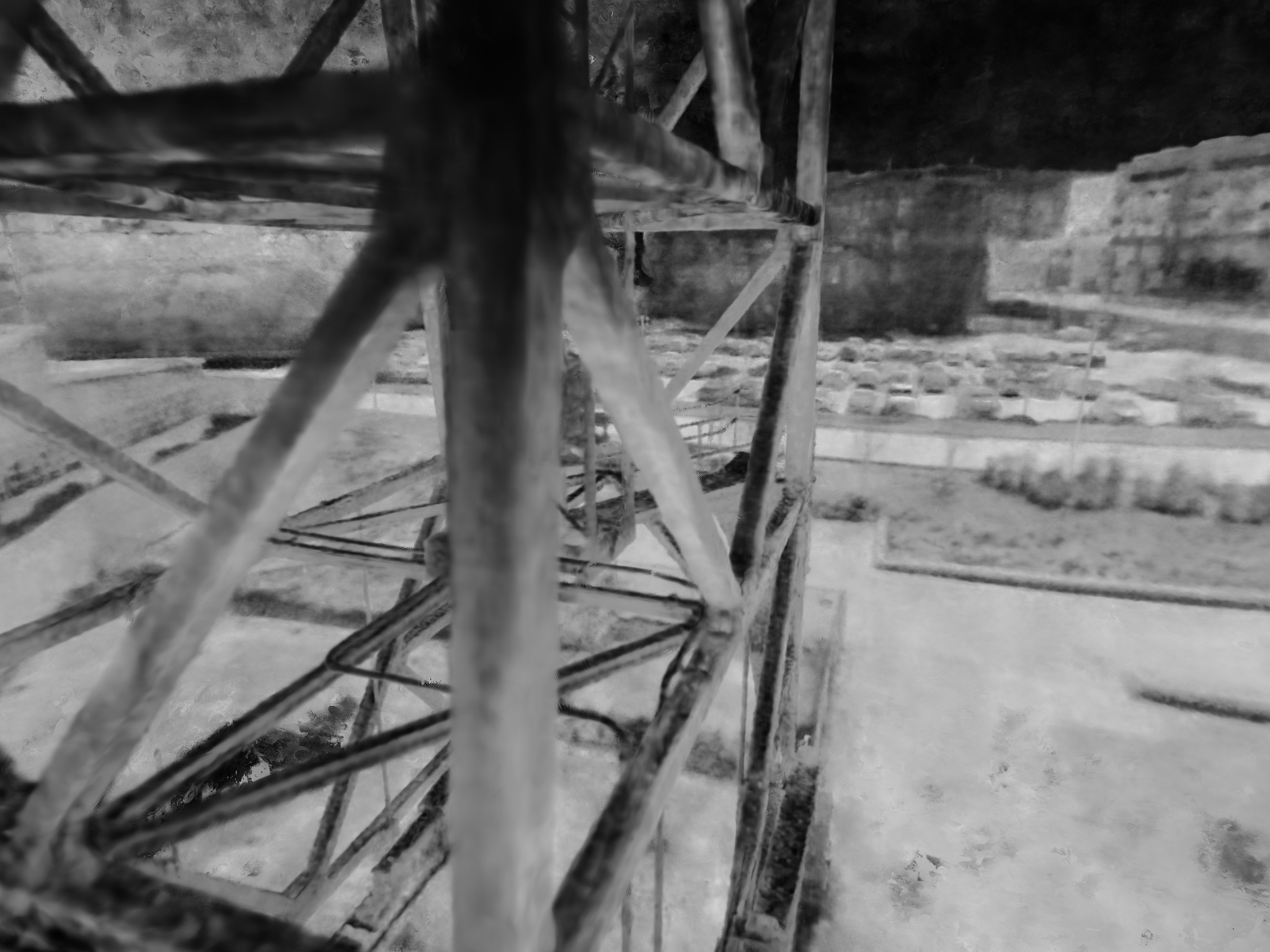}
    \includegraphics[height=0.23\linewidth]{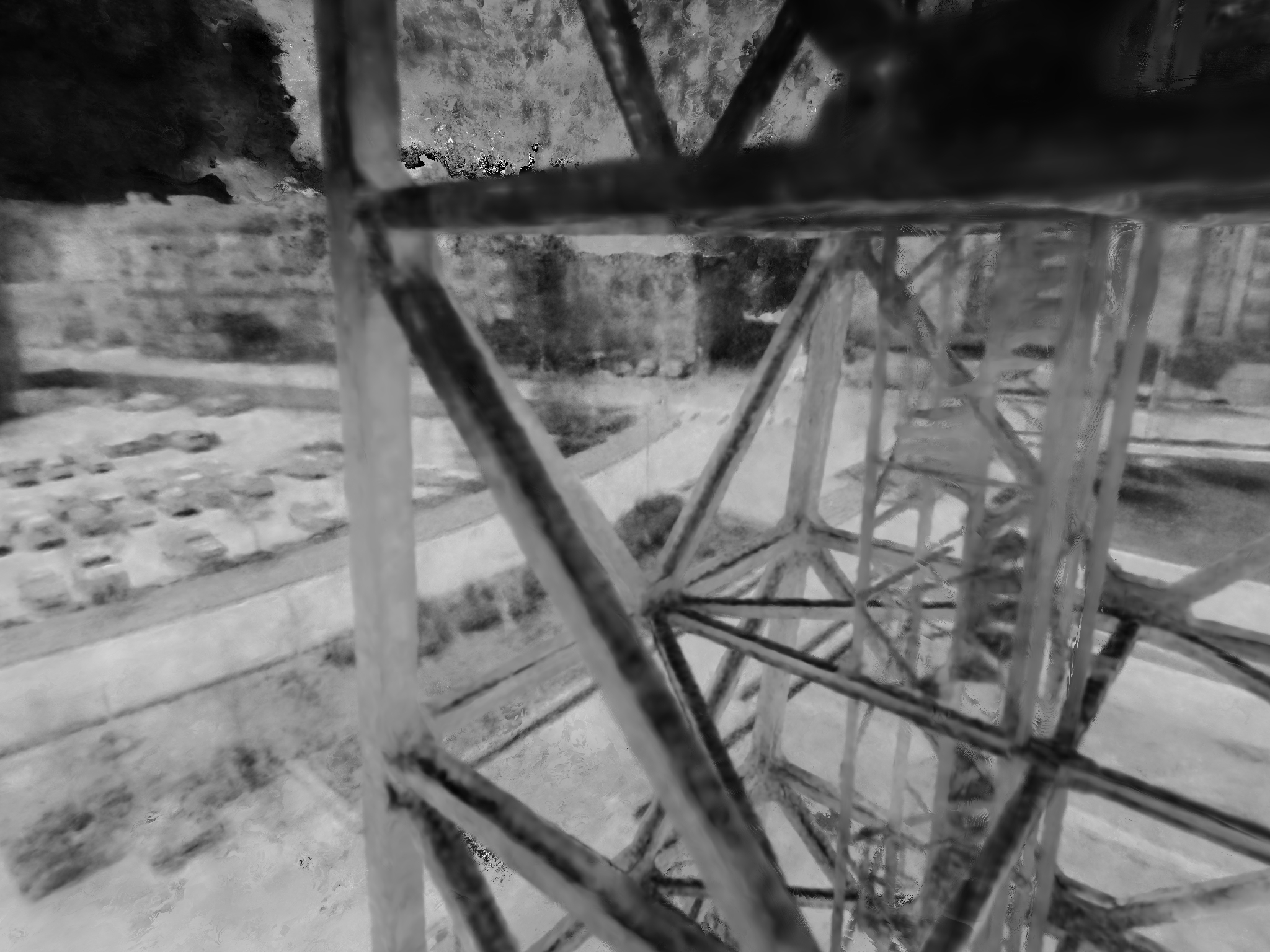}
    \includegraphics[height=0.23\linewidth]{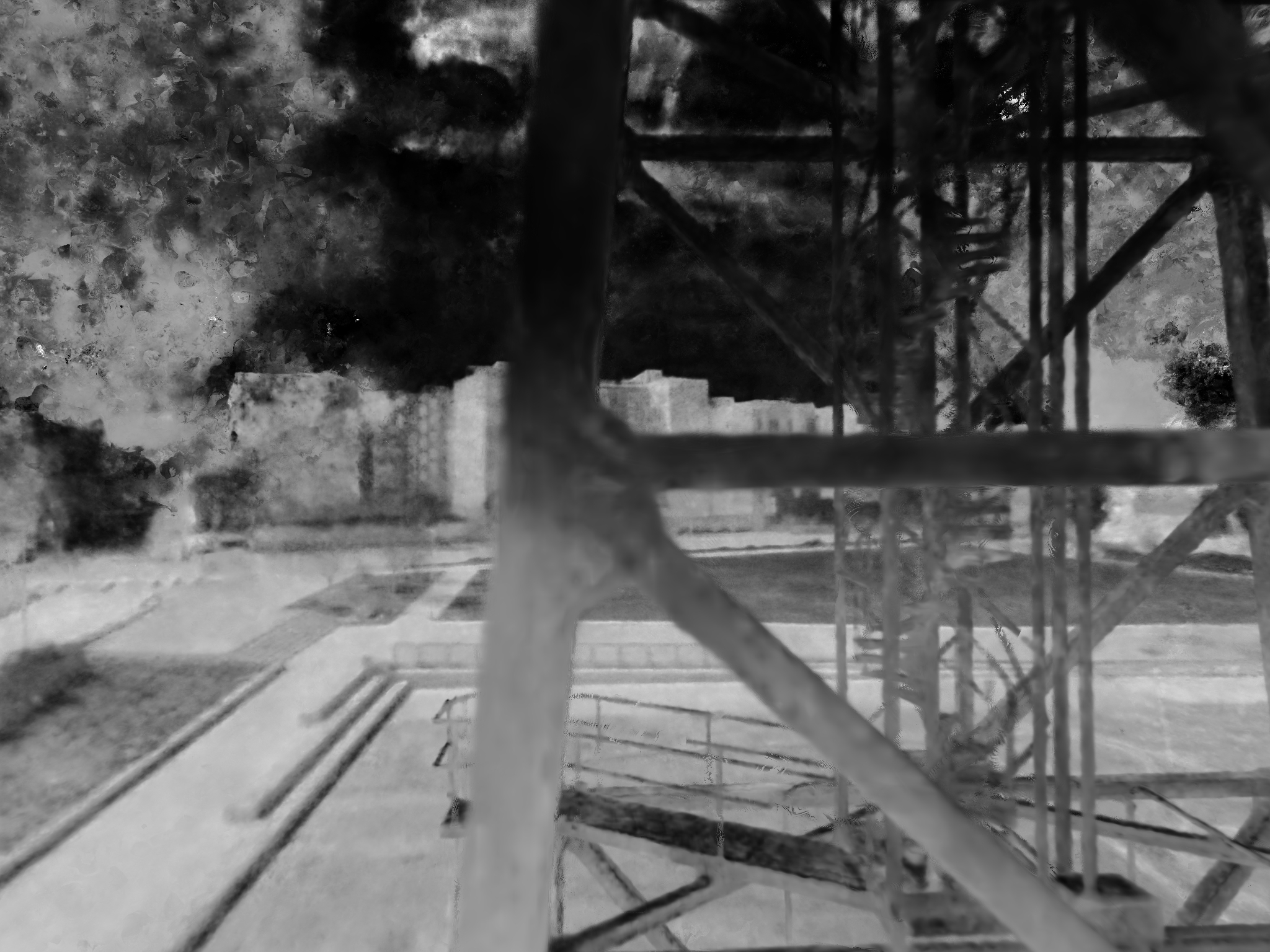}
    \caption{RGB (top) and thermal (bottom) renderings from our reconstruction of a large crane structure, based on images collected by a Skydio drone. Thermal imaging with 3D reconstruction can be used for building and infrastructure inspection, among many other applications ranging from agriculture to search and rescue.}
    \label{fig:skydio}
\end{figure}

\plotscenecropped{pyrex}{00023}{Results on our \emph{pyrex} scene, compared against two Nerfacto-based baselines \cite{nerfstudio} in which the visible and thermal models share a density field \{RGBT\} or are completely separate \{RGB\}\{T\}. The \{RGBT\} model fails to recover the glass bowl as visually transparent yet thermally opaque, while the camera pose refinement fails for the thermal component of the \{RGB\}\{T\} model, leading to ghosting.}

3D radiance field reconstruction has made great strides based on RGB images taken with visible light cameras. However, 3D reconstruction from thermal (long-wave IR) images remains challenging due to the low resolution of thermal cameras. In addition to limiting image quality, this low thermal resolution limits the number of robust 2D image features that can be used to recover thermal camera poses via structure from motion algorithms like COLMAP \cite{schoenberger2016sfm, schoenberger2016mvs} (see \cref{fig:colmap_fail}). Further, even with known thermal camera poses, directly extending existing radiance field models to include a thermal channel produces limited quality 3D reconstructions because many materials interact differently with thermal versus visible light. For example, a glass of water is transparent to visible light but thermally opaque, while smoke is visibly opaque but thermally transparent. 

We propose strategies to address both of these challenges, recovering accurate thermal camera poses by calibrating the relative poses of a thermal camera and an RGB camera, and gracefully combining information from the two spectra while recovering material-specific properties.
Our method also improves 3D thermal reconstruction quality by leveraging information from the visible spectrum for super-resolution, as RGB cameras are often of far higher spatial resolution compared to thermal cameras.
Concretely, we make the following contributions:
\begin{itemize}
    \item We introduce the first method to demonstrate 3D thermal scene reconstruction from long-wave IR images, including cross-calibrating handheld RGB and thermal cameras to estimate thermal camera poses, as well as thermal super-resolution based on fusion of multiview thermal and RGB measurements.
    \item We extend radiance field models to separately represent absorption of thermal and visible light, with appropriate cross-spectral regularization, to enable recovery of material properties and improve reconstruction quality.
    \item We showcase our method on a novel dataset of diverse materials imaged with multiview thermal and RGB cameras, which is available at \url{https://yvette256.github.io/thermalnerf/}. Our dataset includes nine real-world scenes as well as one synthetic scene. We also exhibit our method on a Skydio drone dataset of a large crane structure (see \cref{fig:skydio}).
\end{itemize}

\newcommand{\plottrim}[1]{%
  \adjincludegraphics[trim={{0.13\width} {0\height} {0.13\width} {0\height}}, clip, height=0.35\linewidth]{#1}%
}

\begin{figure}
    \centering
    \fbox{\includegraphics[height=0.35\linewidth]{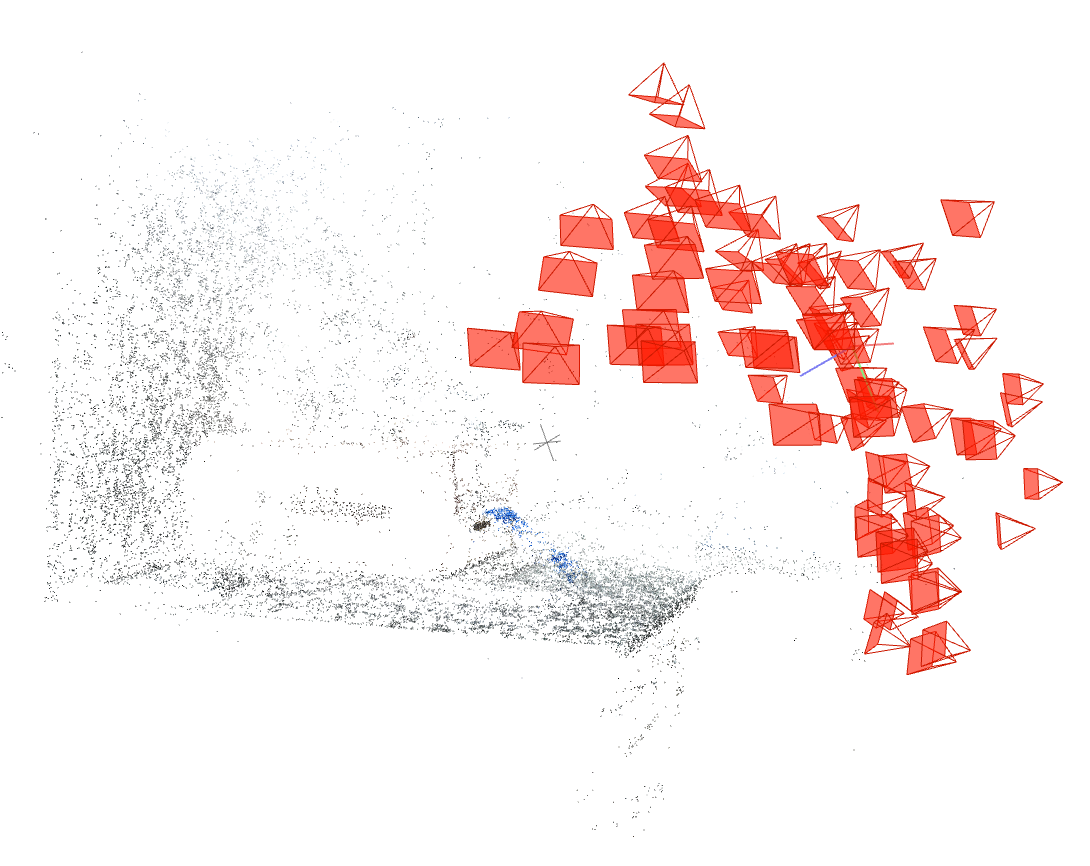}}
    \fbox{\plottrim{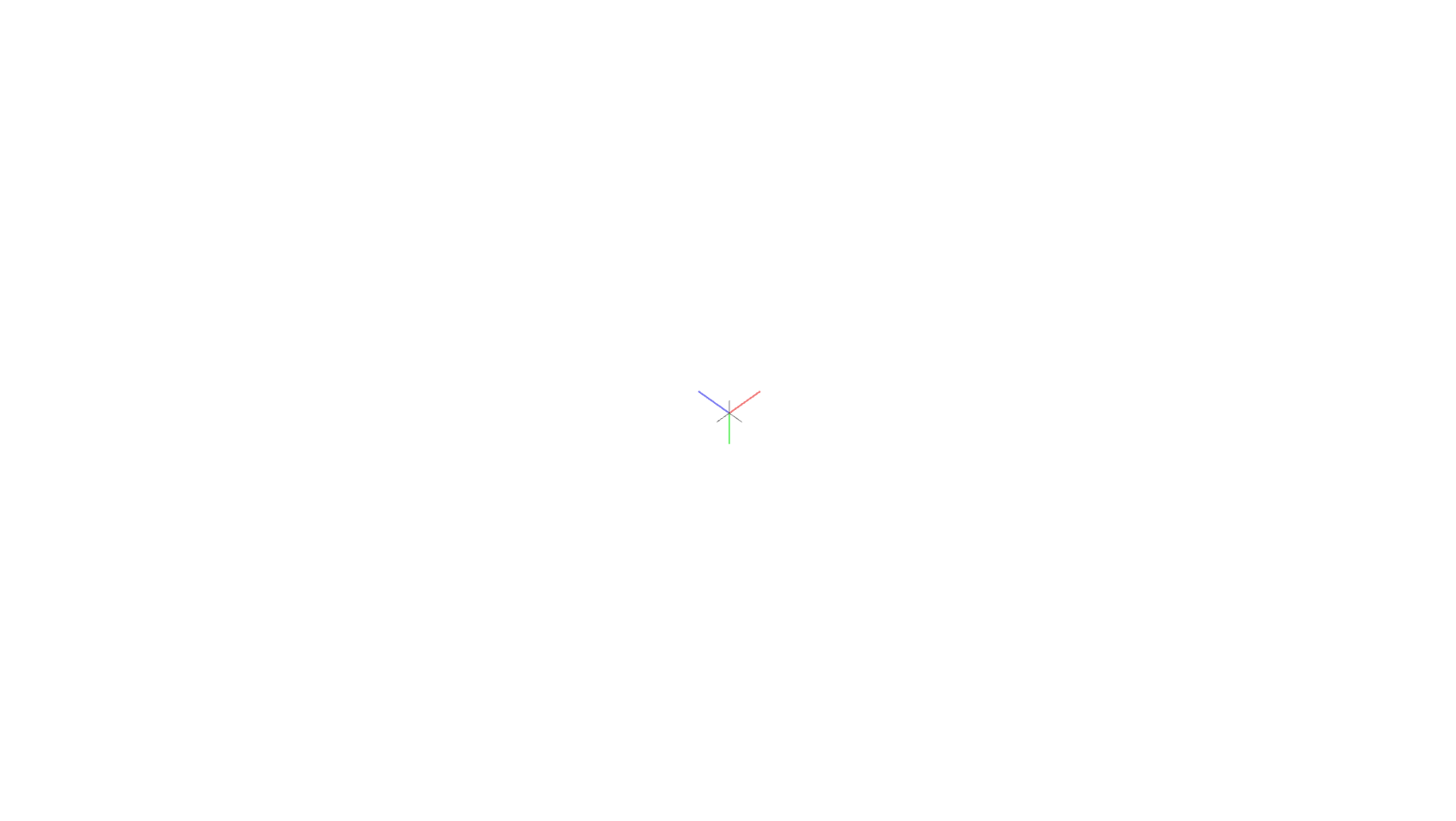}}
    \includegraphics[width=0.233\linewidth]{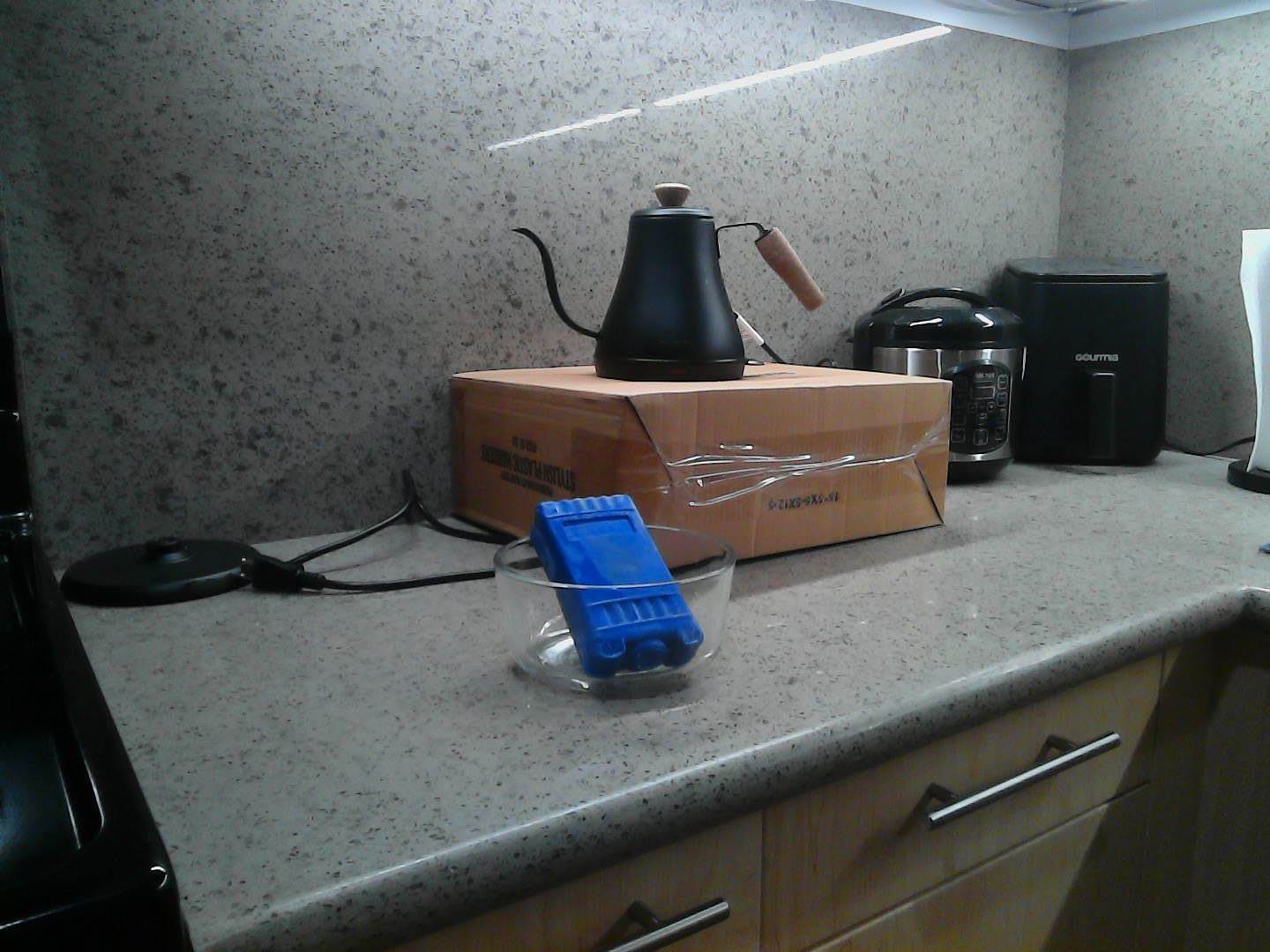}
    \includegraphics[width=0.233\linewidth]{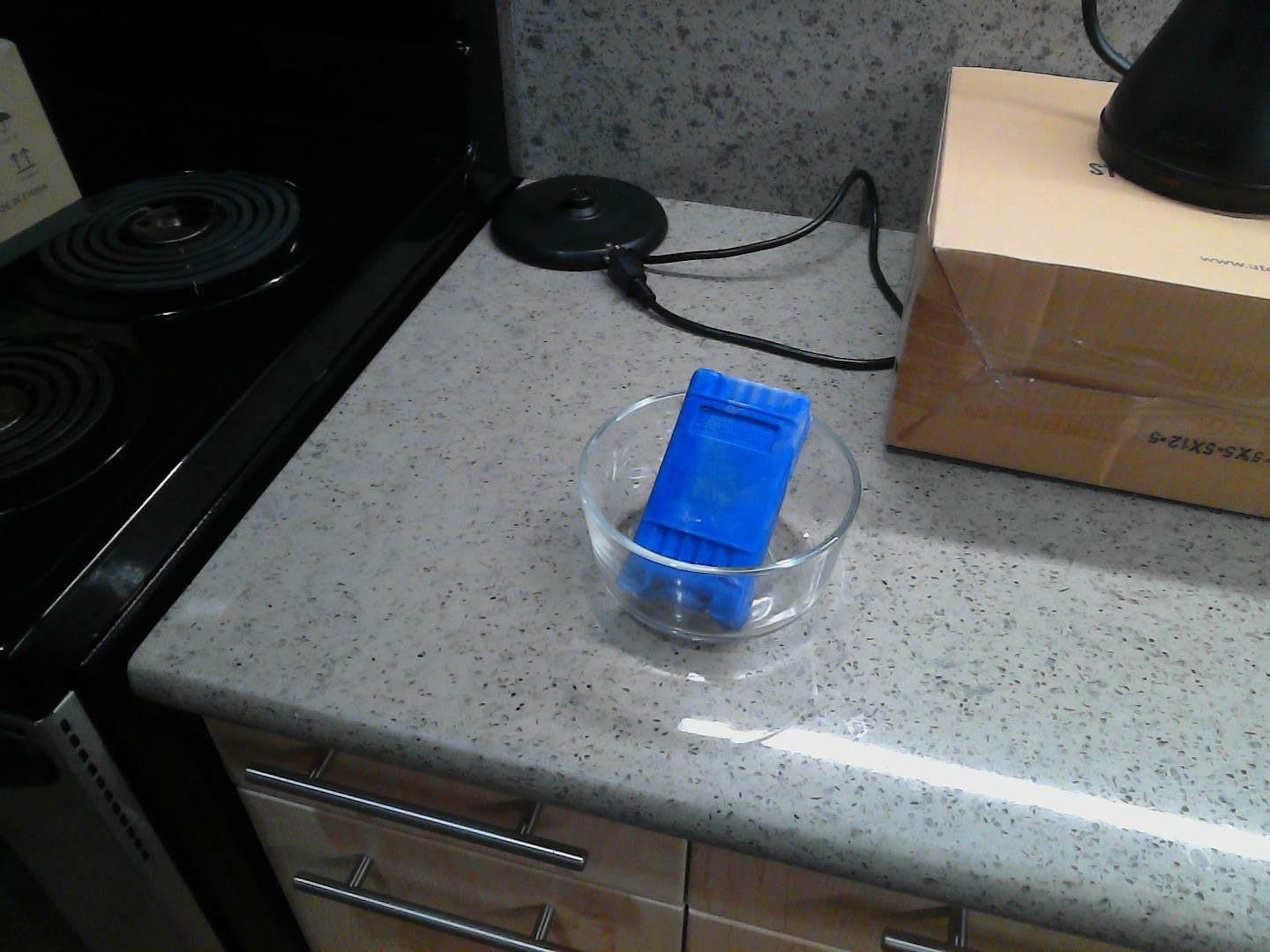}
    \includegraphics[width=0.233\linewidth]{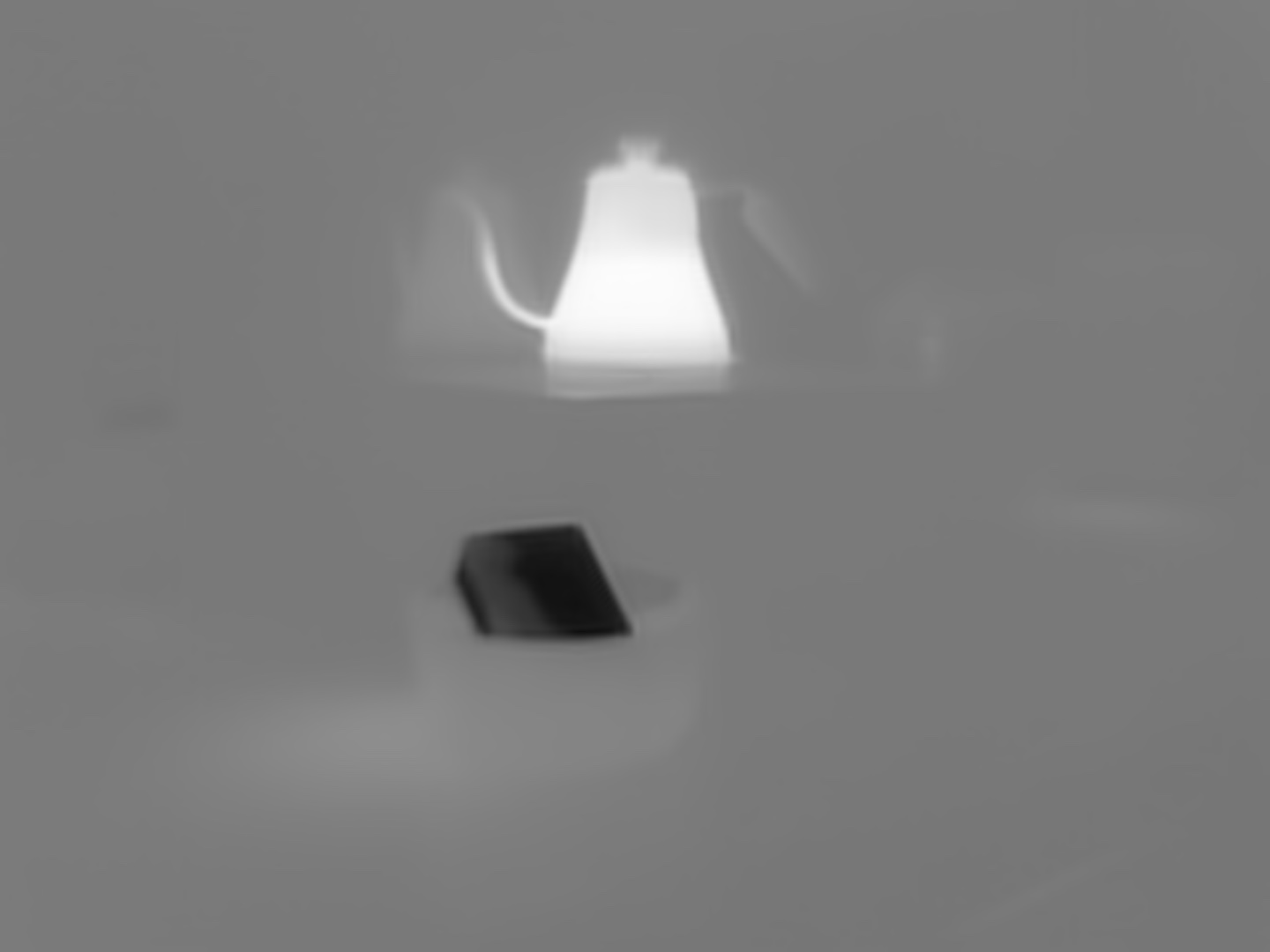}
    \includegraphics[width=0.233\linewidth]{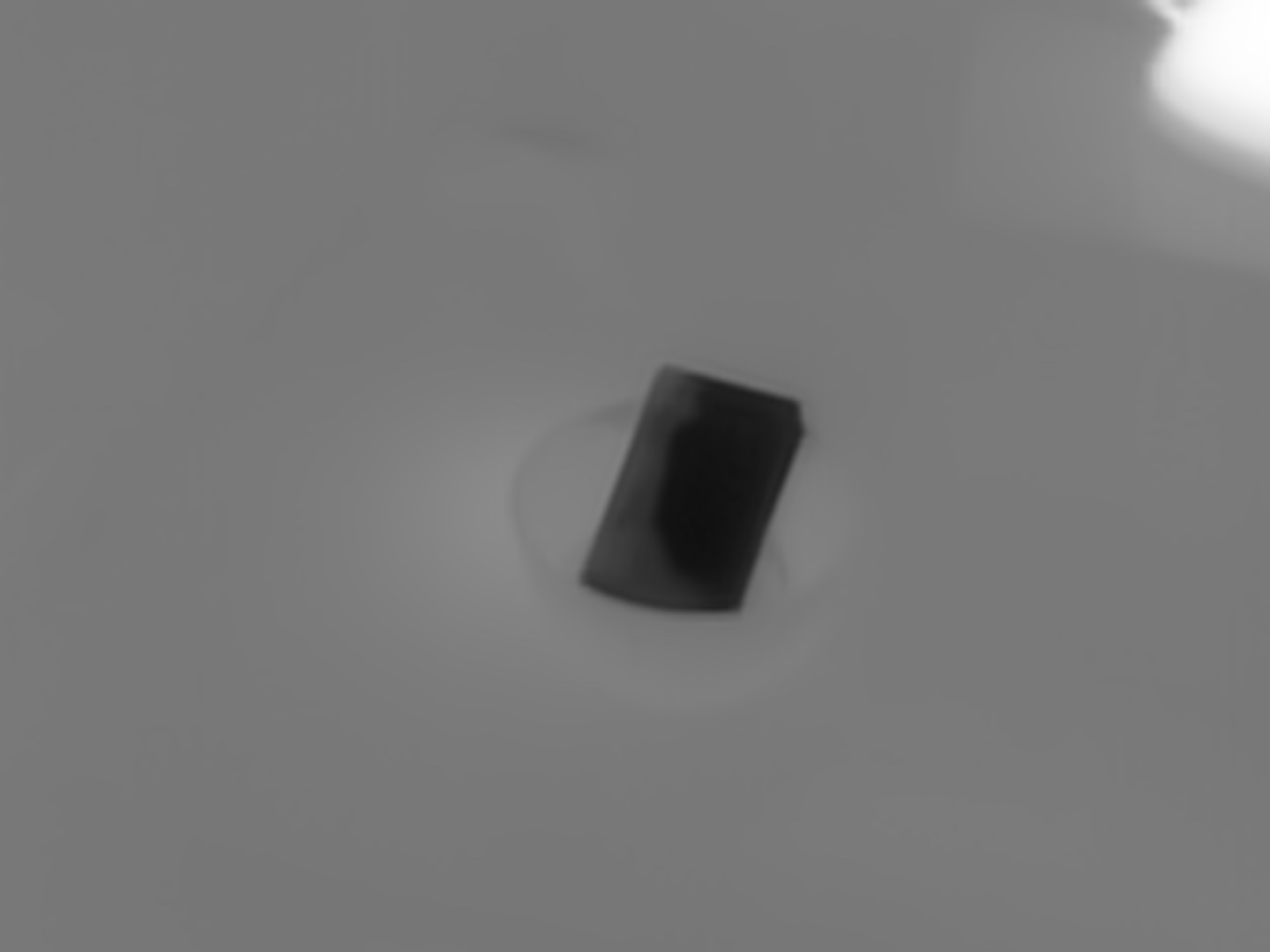}
    \caption{Left: COLMAP successfully recovers camera poses and 3D feature locations from RGB images of our \emph{pyrex} scene. Right: COLMAP fails on corresponding LWIR images of the same scene, due to insufficient features. Bottom: example input views for RGB and LWIR.}
    \label{fig:colmap_fail}
\end{figure}

\section{Related Work}
\label{sec:related}

\subsection{Thermal Imaging}

Thermal cameras detect and measure the heat signature of objects, and convert this emitted infrared (IR) energy into a thermal image reflecting varying levels of IR radiation \cite{flukethermal}. Thermal images can provide insights into scenes and objects that are invisible to visible-light cameras \cite{Fluke_2024}. The contactless nature of thermal imaging adds to its attractiveness in a diverse range of applications, including security and surveillance \cite{wong2009effective}, preventive maintenance \cite{jackson1998leak, concrete_testing}, building inspection \cite{adan2020towards}, monitoring rock masses \cite{grechi20213d}, and archaeology \cite{casana2017archaeological, brooke2018thermal}.

Thermal cameras that measure long-wave infrared (LWIR) wavelengths (8--14 $\mu$m) use very expensive germanium lenses that transmit in the IR spectrum but block visible light \cite{nilsson2008intelligent, kintronics-ip}. The longer wavelength range also requires each element in the detector array to be larger than those required for the visible spectrum \cite{flir-thermal}. These factors contribute to the significantly lower spatial resolution \cite{nilsson2008intelligent, liu2022single} and higher production cost \cite{fluke-blend} of thermal cameras  compared to visible-light cameras. 
Considering this limited resolution, thermal cameras are often paired with other sensors to improve the effective resolution beyond what the thermal camera alone can provide \cite{flir, flirone}.

In particular, 3D scene reconstruction from IR images is a compelling method to achieve such sensor fusion and enhanced resolution, alongside a volumetric scene representation that is of independent value in many applications. Prior work has proposed methods for 3D IR reconstruction by combining short-wave or near IR and visible light images \cite{poggi2022cross}, but these methods yield limited-quality reconstructions and are presently limited to the short-wave IR wavelengths below 1 $\mu$m (see \cref{multispectral}), for which the resolution challenge is far less severe compared to the 8--14 $\mu$m LWIR range we consider. Other approaches have also been proposed for 3D thermal imaging, but these either rely on contact maps from functional grasping \cite{brahmbhatt2019contactdb}, rather than non-contact camera measurements, or study non-line-of-sight object localization rather than direct thermal imaging \cite{maeda2019thermal}. In contrast, our focus is to produce high-fidelity 3D thermal reconstructions of diverse scenes by combining measurements from a low-resolution LWIR camera and a higher-resolution RGB visible light camera, without any direct scene contact.

\plotscene{engine}{00113}{Results on our \emph{engine} scene. The engine is warm from recent driving; thermal imaging is used by mechanics to quickly diagnose problems.}

\subsection{3D Reconstruction and Novel-View Synthesis}

3D reconstruction is a valuable aid in the visualization, survey, and analysis of large or inaccessible objects or landscapes, in ways that 2D images are unable to provide \cite{collaro2025research, jurado2022remote, collaro2023detection}. Broadly, 3D reconstruction can be done based on Structure from Motion (SfM) algorithms using point clouds, or based on fully volumetric models such as radiance fields or signed distance fields. 

Structure from Motion (SfM) algorithms, such as COLMAP \cite{schoenberger2016sfm}, recover both camera poses and 3D points by matching corresponding feature points in 2D images from different viewpoints \cite{robertson2009practical}. After camera poses are known, multi-view stereo (MVS) can then recover a denser 3D reconstruction by matching points across calibrated images \cite{schoenberger2016mvs}. Although MVS often works well, its quality is limited by the accuracy of the camera parameters computed by SfM, as well as reconstruction assumptions such as scene rigidity \cite{furukawa2015multi, wang2024ct}. In particular, COLMAP is used as a preprocessing step in nearly all radiance field models, as a way to recover camera parameters \cite{schoenberger2016mvs, schoenberger2016sfm}. It is usually successful on natural scenes with sufficient high-quality input images, but tends to fail on scenes with limited texture (yielding limited keypoints) or images of low-resolution \cite{cheng2023lu}.

Our work focuses on reconstructing a radiance field as a dense representation of a scene that leverages differentiable rendering instead of feature mapping. Many methods have been proposed for parameterizing a radiance field, including multilayer perceptrons \cite{mildenhall2021nerf,sitzmann2020implicit}, voxel grids \cite{sitzmann2019deepvoxels,lombardi2019neural,yu_and_fridovichkeil2021plenoxels}, factorized tensors \cite{chan2022efficient,chen2022tensorf}, multi-resolution hash tables \cite{mueller2022instant}, and anisotropic 3D points \cite{kerbl3Dgaussians}; we build our method on top of one such representation, the nerfacto model provided by NeRFStudio \cite{nerfstudio}.
While these methods represent great progress in producing faithful and high-resolution radiance fields, they are all focused on the visible spectrum for which high-resolution visible light camera images can be easily acquired. 
Extending radiance field reconstruction to represent other parts of the spectrum which are invisible to the human eye would expose useful and otherwise inaccessible information \cite{zhu2023multimodal}. 

\subsection{Multispectral Radiance Fields} \label{multispectral}
Capturing data beyond the visible spectrum can be helpful in identifying features that are invisible to the usual RGB color channels \cite{zhang2021point, carmona2020assessing, mcleester2018detecting, patrucco20223d, sutherland2023infrared}. 
Accordingly, integrated sensors have been proposed to extend RGB cameras to incorporate information beyond the visible spectrum.

For instance, X-NeRF \cite{poggi2022cross} reconstructs 3D from multispectral images by optimizing a transform between RGB and other sensors including near IR (NIR) cameras \cite{microsoft}, while assuming known camera intrinsics. NIR images tend to have higher resolution as compared to long- and mid-wave IR (LWIR and MWIR) images due to its shorter wavelength \cite{Photonics, Infiniti}. When applied to LWIR images, the novel views rendered from X-NeRF's 3D multispectral reconstruction have relatively low-resolution, suggesting there is room for improved processing of longer wavelength IR images. Additionally, X-NeRF adopts the same assumption as vanilla NeRF: that each material absorbs different wavelengths of light equally, having a shared density parameter regardless of wavelength. While this assumption is a reasonable one for imaging a small range of wavelengths, it becomes problematic for sensor fusion across a wider spectrum like the one we consider.

In addition to these challenges inherent to multispectral imaging, our LWIR spectrum of interest poses additional challenges to 3D reconstruction. LWIR imaging faces an inherent physical resolution limitation due to its longer wavelengths, and
consumer-grade thermal cameras like the handheld FLIR One Pro \cite{flirone} that we use, often have even poorer image resolution \cite{grechi20213d, gonzalez2019thermal}. This poses multiple challenges to obtaining the camera poses and completing the subsequent 3D reconstruction \cite{lopez2021optimized}. We propose a method for thermal 3D reconstruction of scenes using both RGB and thermal images, leveraging cross-camera information to address these challenges \cite{jurado2022remote}. While existing approaches have made use of similar insights in complementary tasks such as dehazing \cite{dumbgen2018near}, hyperspectral imaging \cite{hu2023hyperspectral}, and 3D reconstruction of a person via reflections \cite{liu2023humans}, we demonstrate the first 3D RGB-thermal field reconstruction method that separately models material interactions with thermal and visible spectra to improve reconstruction quality.

\section{Proposed Method}
\label{sec:method}
Our method extends nerfacto \cite{nerfstudio} to the combined RGBT (red, green, blue, and thermal/LWIR) domain. We begin by describing the main idea of our method as compared to standard visible-light RGB radiance fields, and then describing our implementation in more detail.

\plotscene{sheet}{00021}{Results on our \emph{sheet} scene. The plastic sheet is opaque to visible light but transparent to LWIR, revealing the hot kettle behind it.}

\subsection{Main Idea: Broad-Spectrum Radiance Fields}
Existing radiance field models, including NeRF \cite{mildenhall2021nerf} and its many variations, typically focus on modeling radiance in the visible spectrum as three color channels (red, green, and blue). In using the standard volume rendering formula based on the Beer--Lambert law (see \cref{sec:image_formation}), these models implicitly assume that each point in space is equally absorptive of all three of these colors of light. While this is a good approximation for RGB visible light, to which most materials are either opaque or transparent, there are certain materials, such as stained glass, for which the approximation is no longer valid. For example, a red stained glass window transmits red light but occludes green and blue light, violating the equal-absorption assumption.

When we begin to consider radiance fields across a wider spectrum, including our setting of RGB and LWIR thermal radiance field modeling, we find that more materials exhibit differing absorption behavior across this wider spectrum. We model this behavior by explicitly endowing each spatial location with separate densities (absorption coefficients) for each wavelength, while introducing regularization to encourage these wavelength-specific densities to remain similar for most materials.

\subsection{Image Formation Model}
\label{sec:image_formation}

In the standard RGB setting, NeRF represents a scene as
a volumetric radiance field
\[F_\Theta^\rgb: (\vec{x}, \vec{d}) \mapsto (\vec{c}_\rgb, \sigma_\rgb)\]
mapping a 3D point $\vec{x} \in \R^3$ and viewing direction $\vec{d} \in \R^3$
to a volume density $\sigma_\rgb$ and view-dependent emitted color $\vec{c}_\rgb = (r, g, b) \in \R^3$.
The scene is rendered along a camera ray $\vec{r} = \vec{o} + t\vec{d}$ with
origin $\vec{o} \in \R^3$ and direction $\vec{d} \in \R^3$ via standard
volumetric rendering \cite{max1995}
\begin{align}
  \vec{c}_\rgb(\vec{r}) &= \int_0^\infty T(t) \sigma_\rgb(\vec{r}(t)) \vec{c}_\rgb(\vec{r}(t), \vec{d}) dt\\
  \text{where}\quad T(t) &= \exp\left[-\int_0^t \sigma_\rgb(\vec{r}(t')) dt'\right]
\end{align}
which is approximated numerically via $N$ samples along the ray via
\begin{align}
  \vec{c}_\rgb(\vec{r}) &\approx \sum_{i=1}^N w_i \vec{c}_\rgb(\vec{r}(t_i), \vec{d})\\
  \text{where}\quad w_i &= T_i(1 - \exp(-\sigma_\rgb(\vec{r}(t_i))(t_{i+1} - t_i)))\\
  \text{and}\quad T_i &= \exp\left[-\sum_{j=1}^{i-1} \sigma_\rgb(\vec{r}(t_j)) (t_{j+1} - t_j)\right].
\end{align}

Now, to extend NeRF from the RGB to the RGBT domain, given a ray $\vec{r}$, we wish to render
the RGB color $(r,g,b)$ plus the color from a thermal image $\tau$. We treat this
as 4-D color $\vec{c} = (r, g, b, \tau)$. Hence we
introduce $c_\therm := \tau \in \R$:
\begin{align}
  \vec{c}(\vec{r}) = (\vec{c}_\rgb(\vec{r}), c_\therm(\vec{r}))
\end{align}
We observe that in the visible-light spectrum, objects of interest tend to absorb wavelengths of
light similarly, but the same is not true in the combined infrared-and-visible-light spectrum. For example, many objects are opaque to visible light but transparent
to infrared light, or vice versa, as illustrated by the pyrex glass bowl in \cref{fig:pyrex}, which is thermally opaque but visibly transparent. We therefore propose to render $c_\therm(\vec{r})$
with a separate density $\sigma_\therm$, distinct from $\sigma_\rgb$:
\begin{align}
  c_\therm(\vec{r}) &= \int_0^\infty T(t) \sigma_\therm(\vec{r}(t)) c_\therm(\vec{r}(t), \vec{d}) dt\\
  \text{where}\quad T(t) &= \exp\left[-\int_0^t \sigma_\therm(\vec{r}(t')) dt'\right].
\end{align}
We thus represent the RGBT scene as a radiance field
\begin{align}
F_\Theta: (\vec{x}, \vec{d}) \mapsto (\vec{c}_\rgb, c_\therm, \sigma_\rgb, \sigma_\therm).
\end{align}

\plotscene{charger}{00023}{Results on our \emph{charger} scene. Thermal imaging reveals heat dissipated by both the laptop and its power adapter.}

\subsection{Optimization and Regularization}

To optimize $F_\Theta$, we minimize the following objective:
\begin{align}
  \Loss = \Loss_\rgb + \lambda_\therm\Loss_\therm + \lambda_\density \Loss_\density + \lambda_\crosschannel \Loss_\crosschannel + \lambda_\tv \Loss_\tv. \label{eq:loss}
\end{align}

Here $\Loss_\rgb$ and $\Loss_\therm$ are the standard pixel-wise photometric $\ell_2$ losses
against calibrated ground-truth (gt) images:
\begin{align}
    \Loss_\rgb &= \frac{1}{|\mathcal{R}_\rgb|} \sum_{\vec{r} \in \mathcal{R}_\rgb} \left\|\vec{c}_\rgb(\vec{r}) - \vec{c}_\rgb^\text{gt}(\vec{r})\right\|_2^2\\
  \Loss_\therm &= \frac{1}{|\mathcal{R}_\therm|} \sum_{\vec{r} \in \mathcal{R}_\therm} (c_\therm(\vec{r}) - c_\therm^\text{gt}(\vec{r}))^2.
\end{align}
where $\mathcal{R}_\rgb$ and $\mathcal{R}_\therm$ are rays from RGB and thermal cameras respectively.

$\Loss_\density$ is an $\ell_1$ regularizer encouraging the RGB and thermal densities
to deviate from each other only at sparse 3D positions $\vec{x}$: 
\begin{align}
  \Loss_\density &= \frac{1}{|\X|} \sum_{\vec{x} \in \X} \left|\density_\rgb(\vec{x}) - \density_\therm(\vec{x})\right|.
\end{align}
In practice, we implement this regularizer in two parts, where one part applies the regularizer with weight $\lambda_{\density,\rgb}$ to $\density_\rgb$ (while blocking any gradients to $\density_\therm$) and the other applies the same regularizer but with a larger weight $\lambda_{\density,\therm}$ to $\density_\therm$ (while blocking any gradients to $\sigma_\rgb$). This two-part implementation of $\Loss_\density$ allows us to prioritize transfer of information from the RGB to the thermal components of the reconstruction, while still allowing each spectrum to regularize the other.
This two-part $\Loss_\sigma$ regularizer is motivated by the observation illustrated in \cref{fig:pyrex} that, while some materials (like the glass bowl) do absorb visible and infrared light differently, most materials are similarly absorptive of both visible and infrared wavelengths.
For those objects, we can leverage the information in the higher-resolution RGB measurements
to guide the thermal reconstruction and achieve thermal super-resolution.

$\Loss_\crosschannel$ is a variation on the cross-channel prior \cite{crosschanneloriginal, crosschannel2} adapted to our radiance field reconstruction task:
\begin{align}
    \Loss_\crosschannel &= \frac{1}{|\mathcal{R}_\rgb|} \sum_{\vec{r} \in \mathcal{R}_\rgb} \left\|\frac{1}{3}\nabla_\vec{r} (r^\gt + g^\gt + b^\gt)(\vec{r}) - \nabla_\vec{r} c_\therm(\vec{r})\right\|_1.
\end{align}
where $\mathbf{c}^\gt_\rgb = (r^\gt, g^\gt, b^\gt)$.
Concretely, we estimate $\Loss_\crosschannel$ stochastically by selecting a batch of patches of pixels during each minibatch gradient update. For each pixel patch, we convolve with a 2D finite difference kernel to compute the local spatial gradient in each channel,
and then penalize the thermal channel for $\ell_1$ deviation in gradient
relative to the visible channels. Note that $\Loss_\crosschannel$ is only applied to rendered thermal views for which we have ground truth RGB (rather than rendered RGB views for which we have ground truth thermal), so that this loss only affects the thermal reconstruction. Intuitively, this loss encourages object edges (high spatial gradient magnitudes) to align between the thermal and visible reconstructions. Since the RGB resolution of the training images is higher than the thermal resolution, this unidirectional cross-channel loss promotes thermal super-resolution, which demonstrate in \cref{fig:super-res}.

$\Loss_\tv$ is a pixelwise total variation regularizer on thermally-unsupervised rendered thermal views:
\begin{align}
    \Loss_\tv = \frac{1}{|\mathcal{R}_\rgb|} \sum_{\vec{r} \in \mathcal{R}_\rgb} \|\nabla_\vec{r} c_\therm(\vec{r})\|_1.
\end{align}
Concretely, we estimate $\Loss_\tv$ stochastically by selecting a batch of pixel patches during each minibatch gradient update and computing the 2D finite differences to estimate the local spatial gradient. We do this only for rendered thermal views for which we do not have ground truth thermal observations (\textit{i.e.} views for which we only have RGB supervision). We motivate the inclusion of this term with two observations. First, thermal photographs tend to exhibit sparse features. Second, thermal cameras tend to have lower field of view (FOV) than RGB cameras \cite{fov}. Hence, especially when rendered from the perspective of an RGB camera, thermal views of the scene can exhibit noisy artifacts at the edges of the image or in the background. The inclusion of $\Loss_\tv$ discourages the appearance of such artifacts and expands the boundaries of the thermal scene, compensating for the typically lower FOV of the thermal camera.

\subsection{Camera Calibration}
We calibrate both RGB and thermal cameras using OpenCV's camera calibration \cite{opencv} on a 2mm thick aluminum sheet with a 4 $\times$ 11 asymmetric grid of circular cutouts. Each circle has a diameter of $15$mm and a center-center distance of $38$mm.
We chill the aluminum sheet in a freezer, then collect a series of simultaneous RGB and thermal calibration photographs. Example such photographs are shown in \cref{fig:calibration}.

\begin{figure}[h]
    \centering
    \begin{tabular}{c@{}c@{}}
         \includegraphics[width=0.495\linewidth]{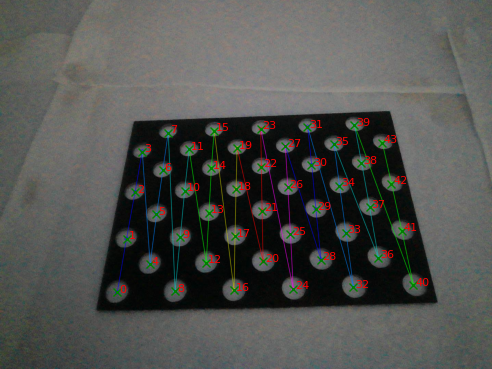} & \includegraphics[width=0.495\linewidth]{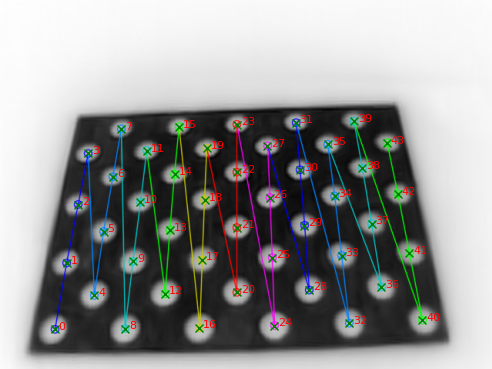} \\
         RGB & Thermal
    \end{tabular}
\caption{Chilled aluminum calibration target as seen simultaneously from our RGB camera (left) and thermal camera (right). Overlaid we show our detected calibration pattern comprised of the $4 \times 11$ detected circles.}
    \label{fig:calibration}
\end{figure}

\plotscene{generator}{00022}{Results on our \emph{generator} scene. Thermal imaging reveals heat emanating from one side of the generator box. Note that the placards on the right side of the box are visible in the \{RGBT\} baseline but not in ours or in the ground truth, since these placards are thermally transparent.}

\plotscene{heater}{00021}{Results on our \emph{heater} scene. The chair in the background is not visible in the ground truth thermal image, or in our reconstruction, but does appear in the \{RGBT\} baseline due to leakage from the RGB channels.}

For each of the RGB and thermal cameras, we estimate
the extrinsic parameters, the rotation $R$ and translation $t$ from the camera coordinates to the world coordinates.
The camera space coordinates $(X_c, Y_c, Z_c)$ corresponding to a point $(X, Y, Z)$ in world space are
\begin{align}
\begin{bmatrix}X_c\\ Y_c\\ Z_c
\end{bmatrix} = \begin{bmatrix}R|t\\ \end{bmatrix}
\begin{bmatrix}X\\ Y\\ Z\\ 1\end{bmatrix}.
\end{align}
We additionally estimate the intrinsic parameters, the focal length $(f_x, f_y)$ and principal point $(c_x, c_y)$, as well as 
the radial distortion coefficients $k_1, k_2$ and the tangential distortion coefficients $p_1, p_2$ of each camera.
Then a pixel $(u, v)$ is determined by 
\begin{align}
\begin{bmatrix}u\\ v\\
\end{bmatrix} = \begin{bmatrix}f_xx'' + c_x\\ f_yy'' + c_y\\ \end{bmatrix}
\end{align}
where
\begin{align}
    \begin{bmatrix}x''\\ y''\\
\end{bmatrix} = \begin{bmatrix}x'(1+k_1r^2+k_2r^4)+2p_1x'y'+p_2(r^2+2x'^2)\\ y'(1+k_1r^2+k_2r^4)+ p_1(r^2+2y'^2)+2p_2x'y'\end{bmatrix}
\end{align}
and
\begin{align}
    \begin{bmatrix}x'\\ y'\\
\end{bmatrix} &= \begin{bmatrix}X_c/Z_c\\ Y_c/Z_c\\ \end{bmatrix}\\
    r^2 &= x'^2 + y'^2.
\end{align}

After using OpenCV's camera calibration \cite{opencv} to solve for the intrinsic and extrinsic parameters of both cameras, we compute the relative transform between the RGB camera and the thermal camera.
We use COLMAP \cite{schoenberger2016sfm, schoenberger2016mvs} to estimate camera poses for our RGB camera for each scene, and then recover thermal camera poses using this calibrated relative pose. We resolve the global scale ambiguity in applying this camera offset by measuring the physical distance between two of our camera measurements for each scene.

\plotscene{trace}{00051}{Results on our \emph{trace} scene. Thermal imaging can locate gaps in insulation, illustrated by the window behind the couch.}

\subsection{Training}
\label{ssec:training}

We build and train our model by building upon the NeRFStudio library \cite{nerfstudio}, using a modification of the default nerfacto model. Specifically, we create a second copy of the entire nerfacto model (including the auxiliary sampling model) to represent the thermal scene, and train the two models in parallel on their respective training images. The only connection between the RGB and LWIR (thermal) models is defined by our regularizers $\Loss_\density$ and $\Loss_\crosschannel$.

\subsection{Dataset Collection}
\label{ssec:dataset}

We test our method on a novel dataset of 9 real-world scenes, each with 50--170 images from distinct viewpoints. We collected these images using a handheld FLIR One Pro \cite{flir}, which attaches to a smartphone and records simultaneous RGB and LWIR channel images at a resolution of 1080 $\times$ 1440 and 480 $\times$ 640, respectively. For each scene, we reserve 10\% of the RGB-LWIR image pairs as a test set, and train on the remaining 90\%. We selected these scenes to demonstrate a range of indoor and outdoor settings for which thermal imaging shows interesting phenomena, such as revealing heat sources and sinks, checking thermal insulation, and imaging through visibly occlusive media. Example images from each scene are shown in \cref{fig:pyrex}, \cref{fig:engine}, \cref{fig:sheet},  \cref{fig:charger}, \cref{fig:generator}, \cref{fig:heater}, \cref{fig:trace}, \cref{fig:sink}, and
\ifsupplement
\cref{fig:generators}
\else
fig. 14
\fi
in the supplement.

In addition to this real-scene dataset, we also introduce a synthetic RGBT scene based on the \emph{hotdog} scene from the NeRF Blender dataset \cite{mildenhall2021nerf}, in which we make the hotdog thermally hot. This synthetic scene allows us to separately test the regularization and modeling aspects of our approach, without any concerns over potential miscalibration. It also allows us to test our thermal super-resolution capability in a setting for which ground-truth high-resolution thermal images are available, which is not the case for our real-scene dataset due to the limited thermal resolution of the FLIR One Pro. 
For this synthetic dataset, we render 45 training views and 5 testing views, each with red, green, blue, and thermal channels.

\section{Experimental Results}
\subsection{Qualitative Results}

We show qualitative results on all of our real-world scenes in \cref{fig:pyrex}, \cref{fig:engine}, \cref{fig:sheet}, \cref{fig:charger}, \cref{fig:generator}, \cref{fig:heater}, \cref{fig:trace}, \cref{fig:sink}, and
\ifsupplement
\cref{fig:generators}
\else
fig. 14
\fi
in the supplement. We note that, while all methods tend to do decently well at reconstructing the RGB scene, only our method reliably also recovers the thermal scene. Treating the thermal and visible spectra separately (\{RGB\}\{T\}) typically results in catastrophic failure to reconstruct the thermal scene, because the low resolution of the thermal images causes the camera pose refinement of NeRFStudio \cite{nerfstudio} to diverge---a similar behavior as in \cref{fig:colmap_fail}. Treating the thermal and visible spectra jointly (\{RGBT\}), with a shared density field, is successful on many objects but fails to reconstruct certain materials, like glass, and often exhibits unintended leakage from the visible to the thermal spectra.

\subsection{Quantitative Results}

In Table~\ref{tab:baselines} we compare our method to two natural baseline approaches based on nerfacto \cite{nerfstudio}. The \{RGB\}\{T\} baseline consists of completely separate nerfacto models for the visible and thermal wavelengths, while the \{RGBT\} baseline consists of a single nerfacto model with an extra color channel to represent the thermal (LWIR) spectrum. Our method is similar to the \{RGB\}\{T\} baseline, but with regularizers to tie together the visible and thermal reconstructions for most materials. 

While both of these baselines successfully recover the RGB scene--unsurprisingly since the nerfacto model was developed for RGB imaging--only our method is also successful at simultaneous thermal reconstruction. The \{RGB\}\{T\} baseline tends to fail dramatically due to diverging thermal camera pose refinement (built into NeRFStudio \cite{nerfstudio}), while the \{RGBT\} baseline suffers thermal artifacts for not allowing any density (absorption coefficient) variation between the visible and thermal reconstructions.

\begin{table*}[!h]
\renewcommand{\arraystretch}{1.3}
\caption{Quantitative results and ablations, comparing our method against nerfacto baselines with fully joint or fully separate models for RGB and thermal spectra, as well as versions of our method without each of our proposed regularizers. The best result in each column is bolded and the second-best is underlined.}
\centering
\begin{tabular}{lcccccc}
\hline
& \multicolumn{2}{c}{PSNR ($\uparrow$)} & \multicolumn{2}{c}{SSIM ($\uparrow$)} & \multicolumn{2}{c}{LPIPS ($\downarrow$)}\\
Method & RGB & Thermal & RGB & Thermal & RGB & Thermal \\
\hline
Ours & 23.26 & \textbf{31.29} & \textbf{0.760} & \textbf{0.945} & 0.387 & \textbf{0.055}  \\
\hline
Nerfacto \{RGBT\} & 21.89 & 21.51 & 0.721 & 0.845 & 0.417 & 0.182\\
Nerfacto \{RGB\}\{T\} & 22.35 &19.23 & 0.726 &0.871& 0.397 &0.226\\
\hline
Ours w/o $\Loss_\crosschannel$ & 23.25 & \underline{30.53} & \underline{0.759} & 0.928 & \underline{0.386} & 0.064\\
Ours w/o $\Loss_\density$ & \underline{23.28} & 29.76 & 0.758 & 0.928 & \textbf{0.382} & 0.062 \\
Ours w/o $\Loss_\tv$ & \textbf{23.33} & 30.52 & \underline{0.759} & \underline{0.931} & \underline{0.386} & \underline{0.059} \\
\hline
\end{tabular}
\label{tab:baselines}
\end{table*}

\subsection{Ablation Studies}

\Cref{tab:baselines} also includes ablation studies comparing our method to versions of it without each of our three regularizers, $\Loss_\crosschannel$, $\Loss_\density$, and $\Loss_\tv$. Note that our method without any regularization is identical to the \{RGB\}\{T\} baseline, except that we also reduce the degree of camera pose refinement.

We find that including any regularizer alone produces a substantial improvement in thermal reconstruction quality relative to the two nerfacto baselines, with modest improvement in RGB reconstruction quality. 
We note that, in addition to the modest quantitative impact of $\Loss_\density$ shown in \cref{tab:baselines}, this regularizer produces meaningful qualitative improvement.
This qualitative improvement can be seen via the RGB and thermal depth maps in \cref{fig:depthmapsablation}, and by enabling the de-occlusion application shown in \cref{fig:removal}.

\begin{figure}
    \centering
    \begin{tabular}{l@{~~}c@{}c@{}}
        \rotatebox{90}{~~~~~~~~~~Ours} & \includegraphics[width=0.466\linewidth]{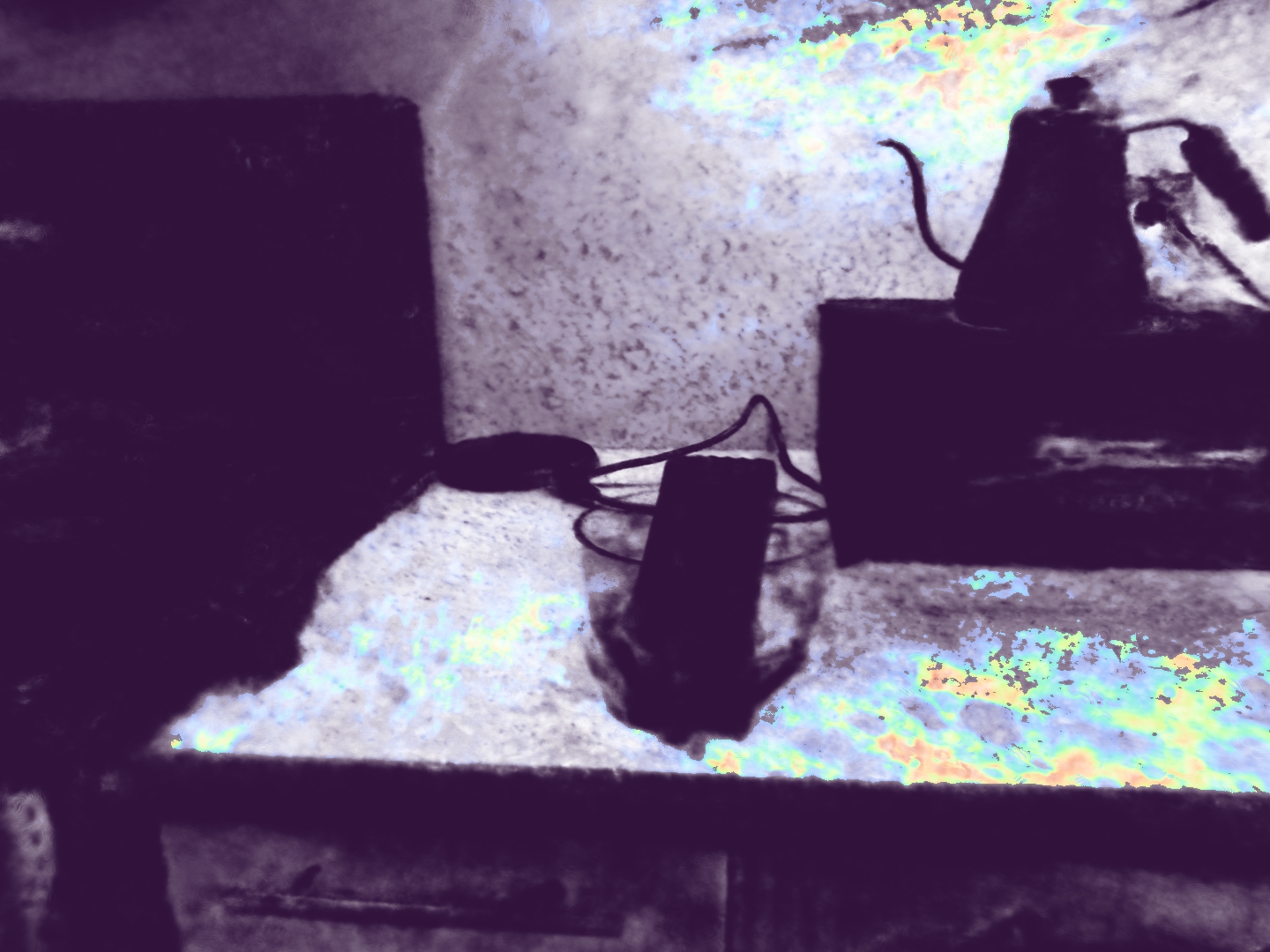} &
        \includegraphics[width=0.466\linewidth]{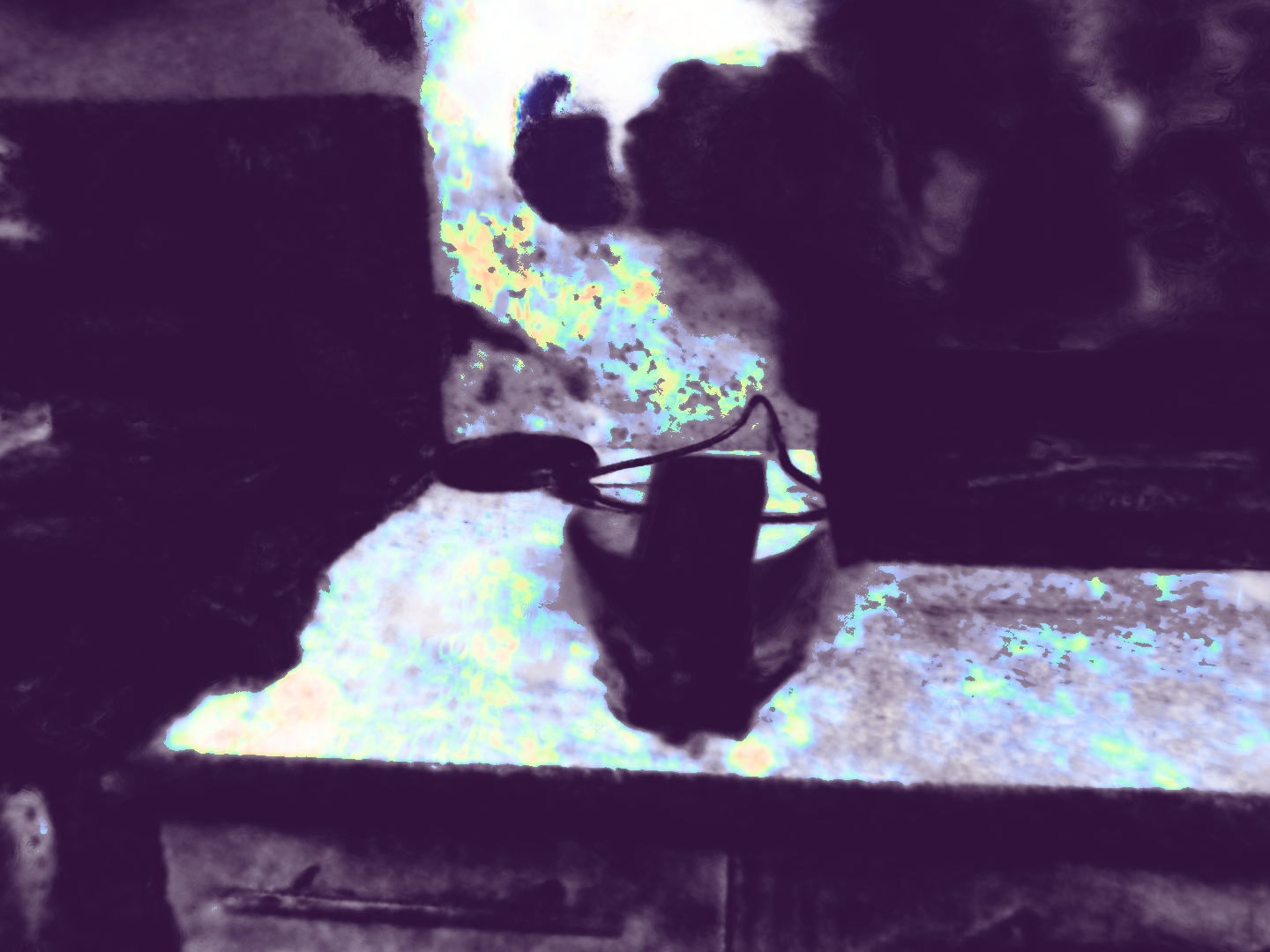}\\
        \rotatebox{90}{~~~~~~~~~~Ours w/o $\Loss_\density$} & \includegraphics[width=0.466\linewidth]{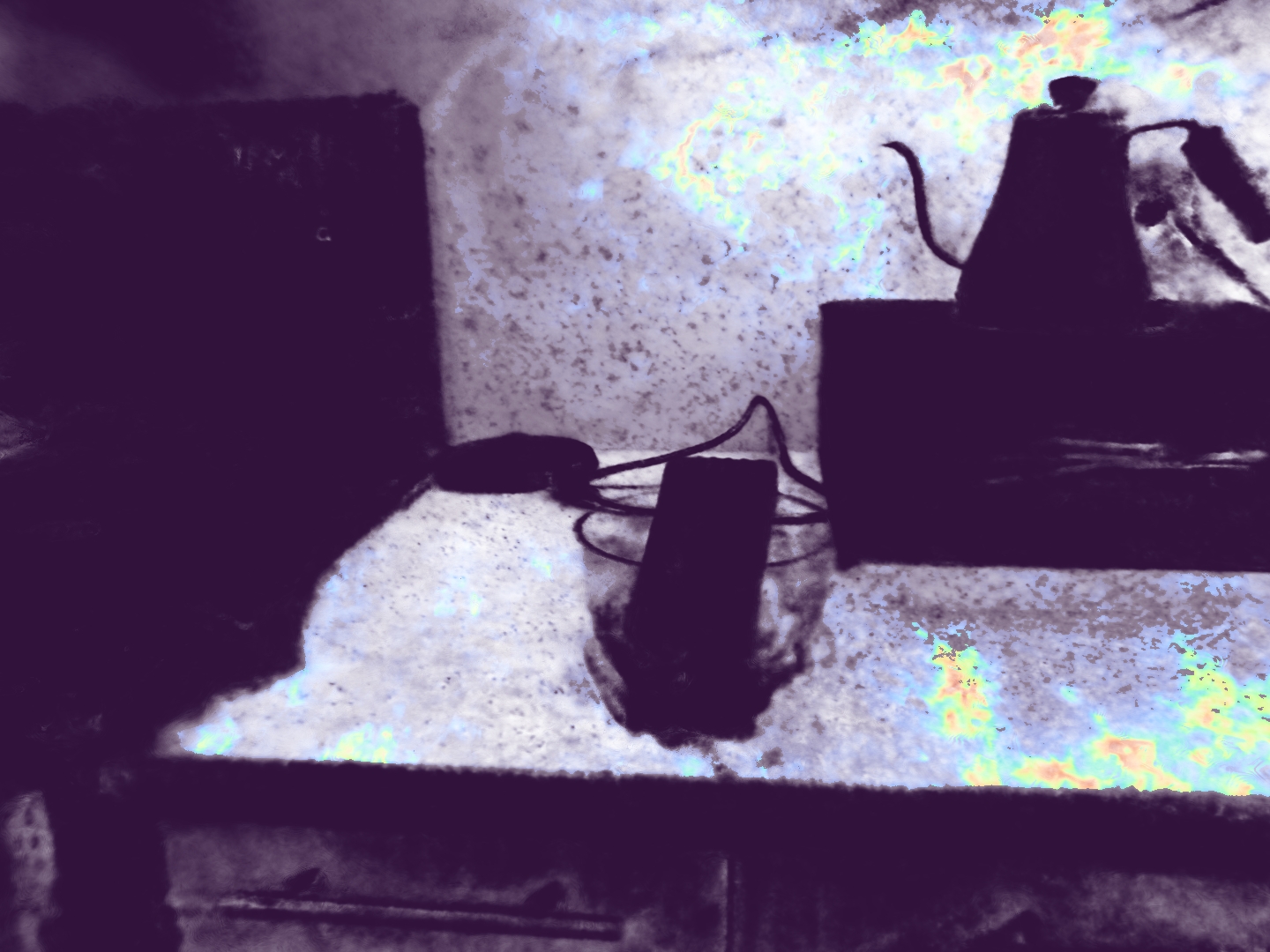} & 
        \includegraphics[width=0.466\linewidth]{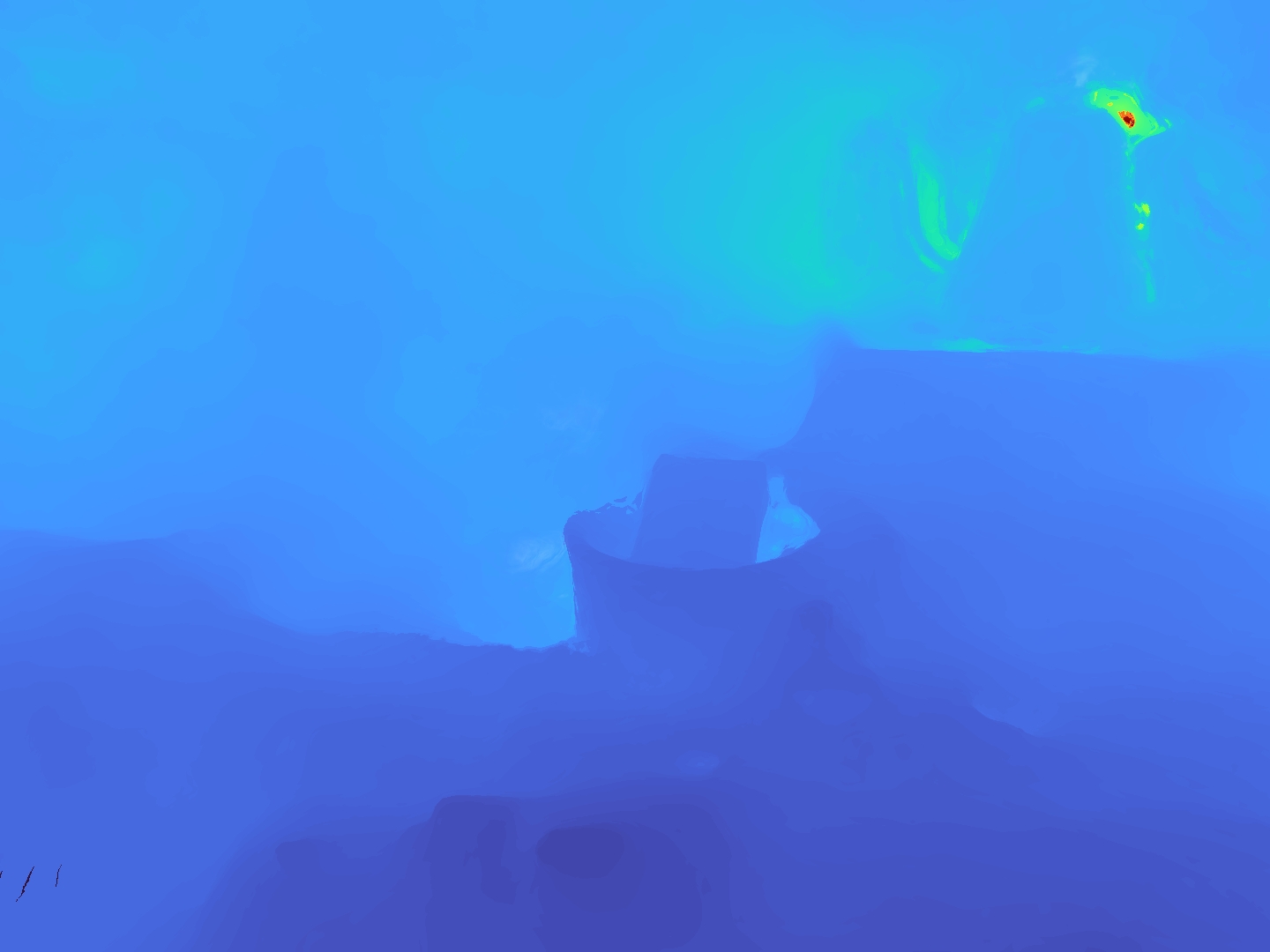}\\
        & RGB Depth & Thermal Depth
    \end{tabular}
    \caption{Depth maps of \emph{pyrex} with and without $\Loss_\density$. Top: RGB and thermal depth maps for our method. $\Loss_\density$ encourages the RGB and thermal densities to resemble each other, resulting in depth maps where the difference between the visible-light and thermal material properties is clear e.g. in the opacity of the glass container. Bottom: RGB and thermal depth maps for our method without $\Loss_\density$. The RGB and thermal densities are not constrained to be similar, resulting in less instructive depth maps.}
    \label{fig:depthmapsablation}
\end{figure}

\begin{figure}[!h]
    \centering
    \begin{tabular}{c@{}c@{}}
         \includegraphics[width=0.495\linewidth]{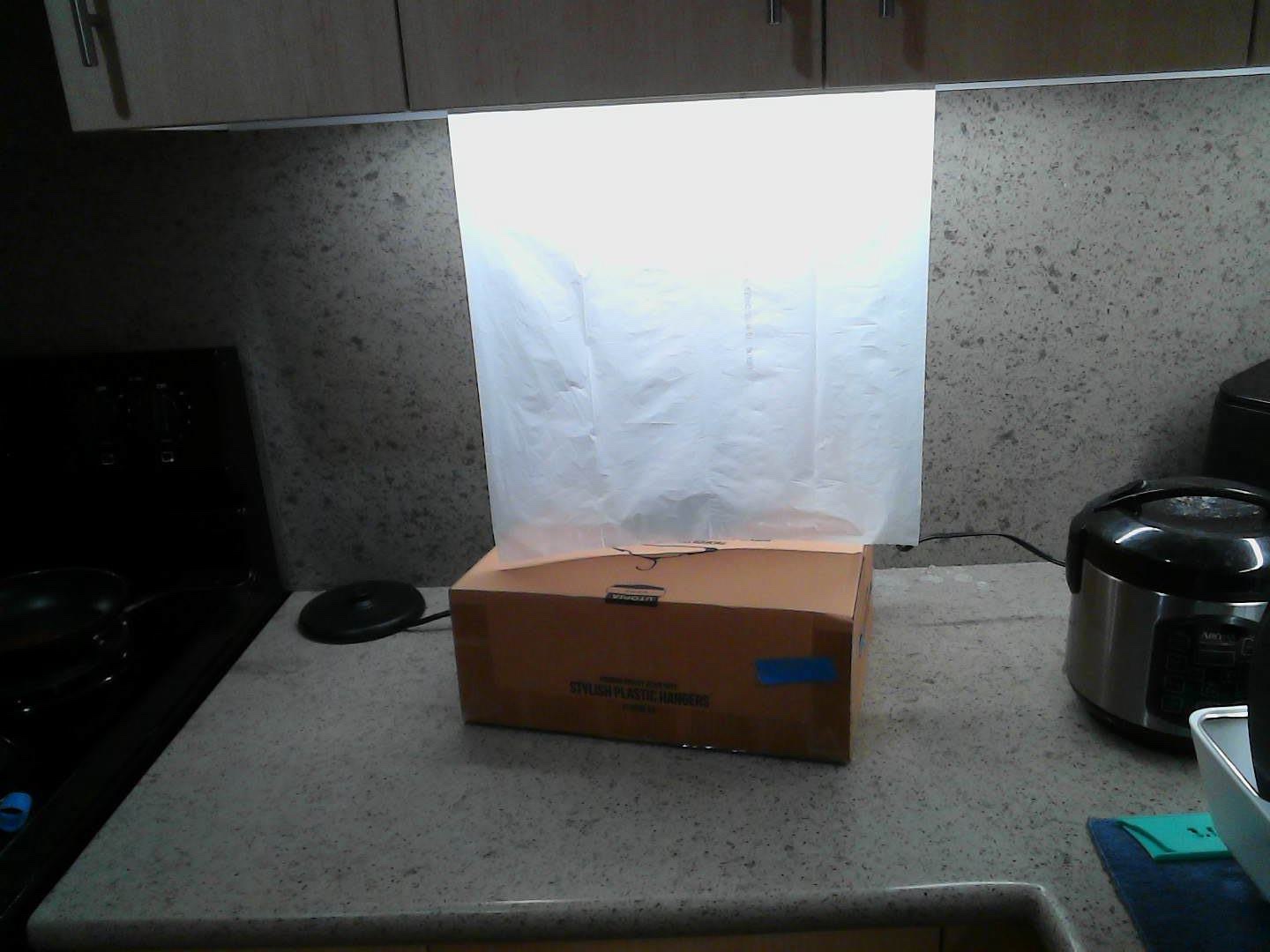} & \includegraphics[width=0.495\linewidth]{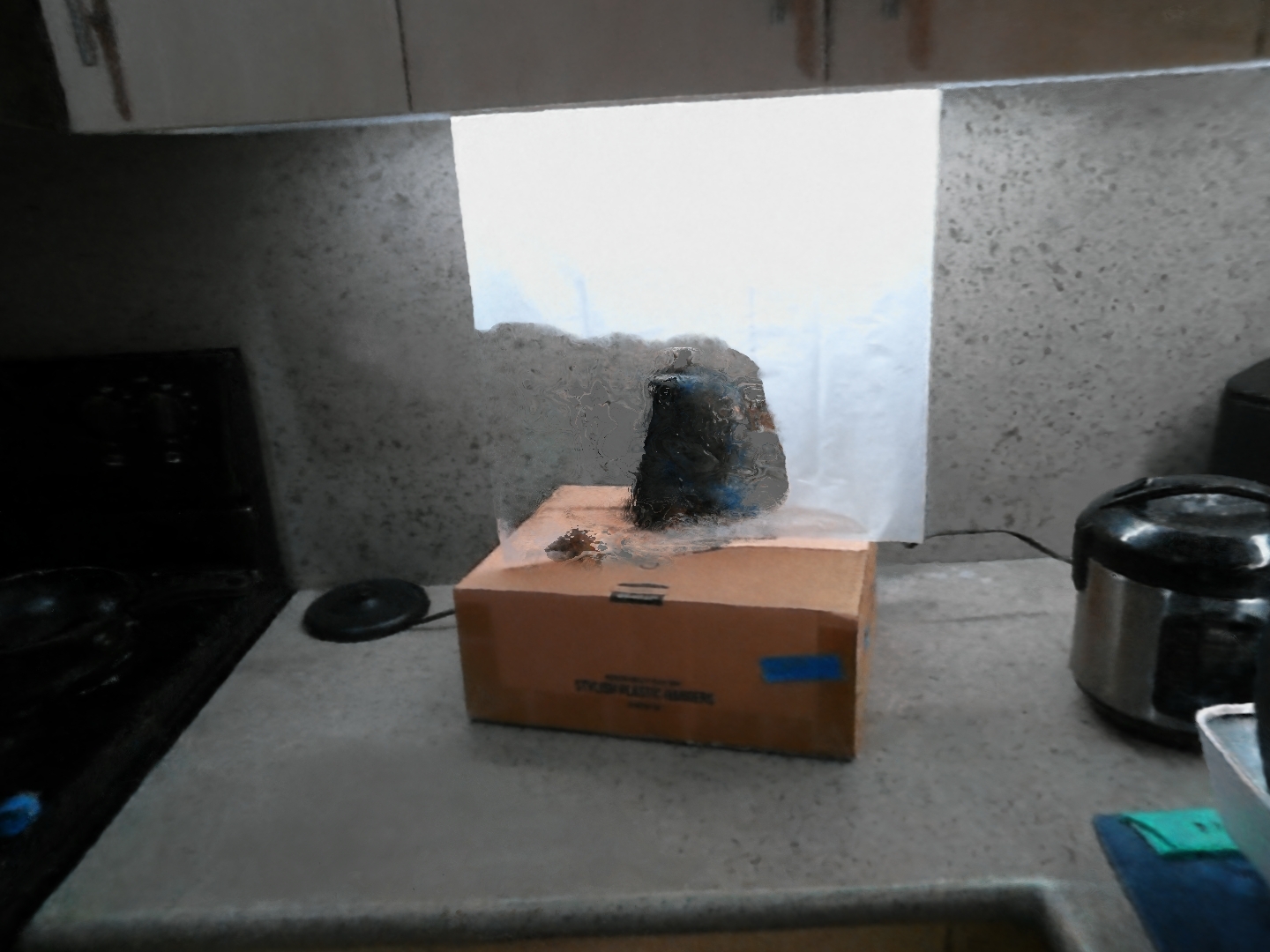} \\
         \includegraphics[width=0.495\linewidth]{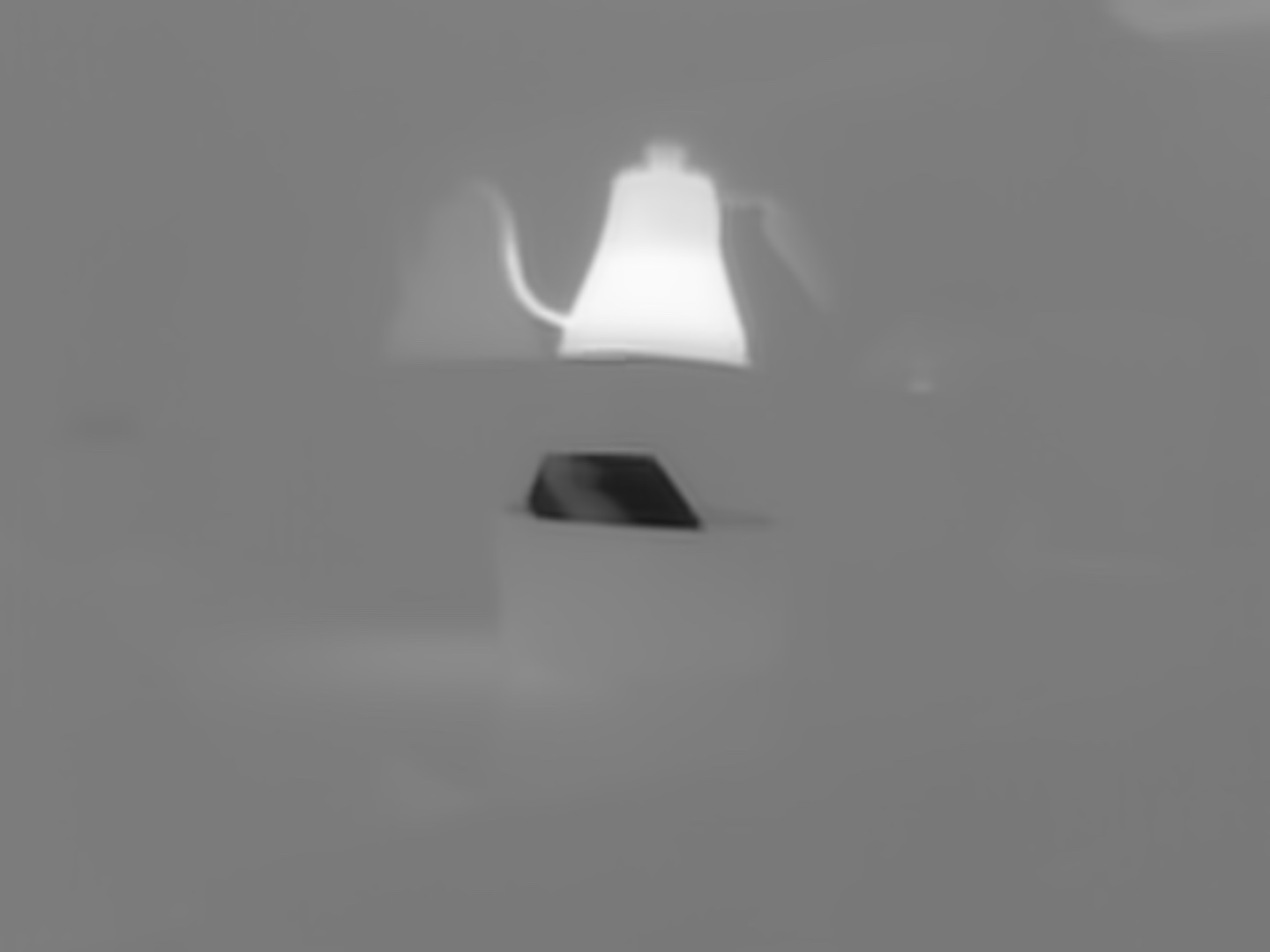} & \includegraphics[width=0.495\linewidth]{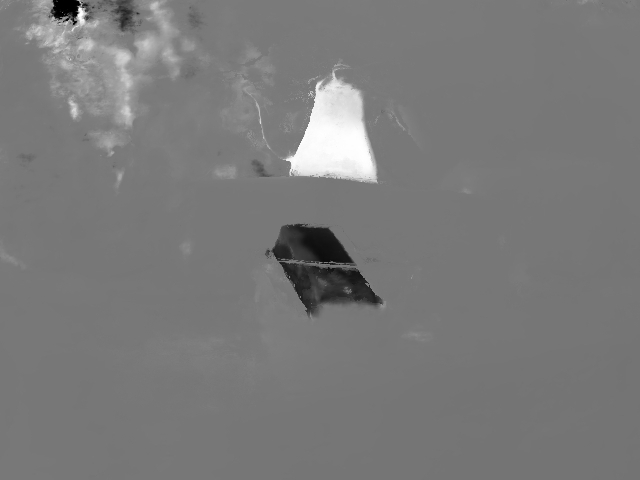}\\
         Ground truth & Hidden object revealed
    \end{tabular}
\caption{Revealing hidden objects. Top: The \emph{sheet} scene rendered without areas of low thermal density, revealing the tea kettle behind the plastic sheet, which is not visible in the ground-truth RGB observation from this viewpoint. Bottom: The \emph{pyrex} scene rendered without areas of low RGB density, revealing the cold pack within the glass container, which is not visible in the ground-truth thermal observation from this viewpoint.}
    \label{fig:removal}
\end{figure}

\begin{figure*}[!h]
\centering
\begin{tabular}{c@{} c@{} c@{} c@{}}
\includegraphics[width=0.23\linewidth]{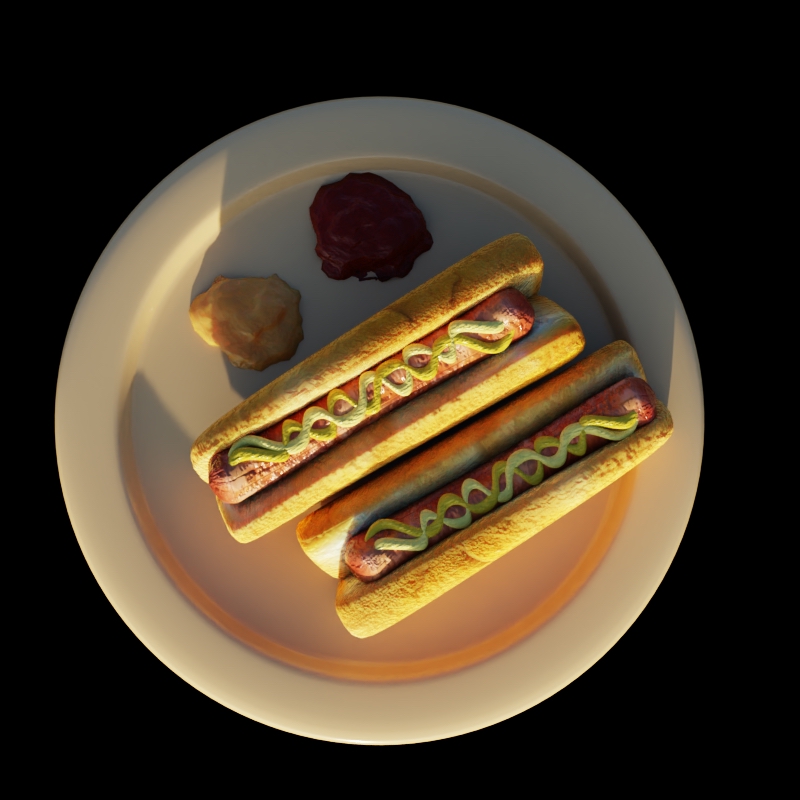}&
\includegraphics[width=0.23\linewidth]{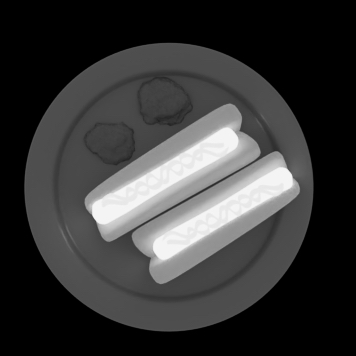}&
\includegraphics[width=0.23\linewidth]{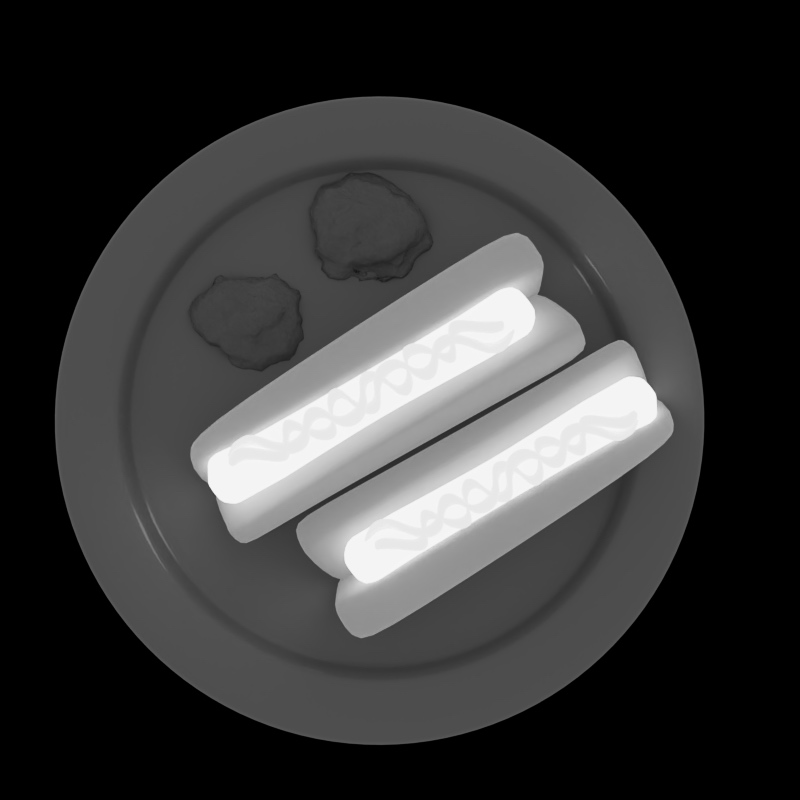}&
\includegraphics[width=0.23\linewidth]{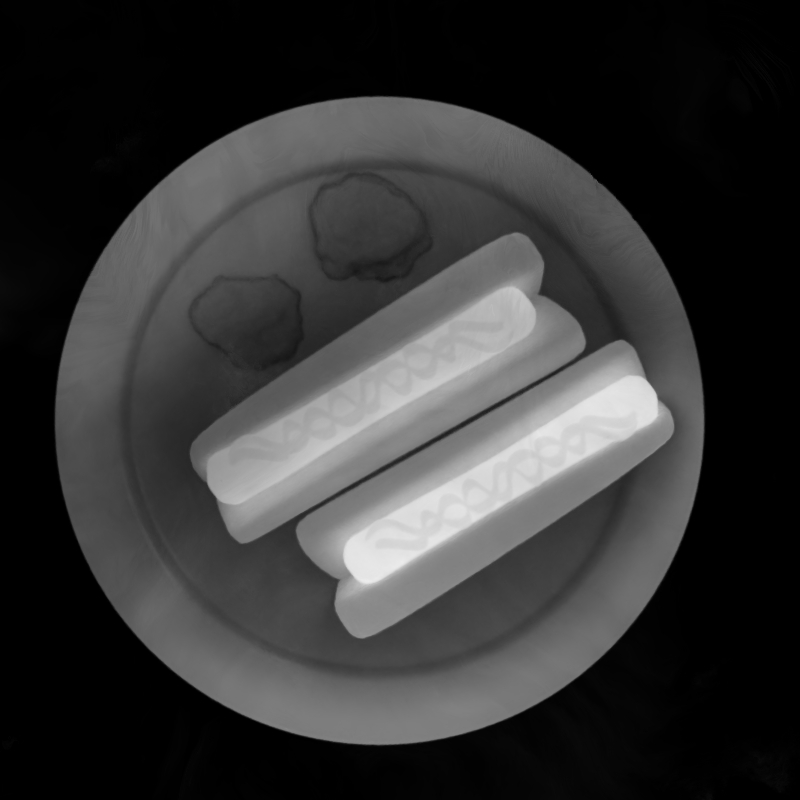}\\
    \adjincludegraphics[trim={{0.75\width} {0.45\height} {0.15\width} {0.45\height}}, clip, width=0.23\linewidth]{figs/hotdog/ground_truth_rgb/r_35.jpg} & 
    \adjincludegraphics[trim={{0.75\width} {0.45\height} {0.15\width} {0.45\height}}, clip, width=0.23\linewidth]{figs/hotdog/ground_truth_thermal/r_35_small.jpg} & 
    \adjincludegraphics[trim={{0.75\width} {0.45\height} {0.15\width} {0.45\height}}, clip, width=0.23\linewidth]{figs/hotdog/ground_truth_thermal/r_35_large.jpg} & 
    \adjincludegraphics[trim={{0.75\width} {0.45\height} {0.15\width} {0.45\height}}, clip, width=0.23\linewidth]{figs/hotdog/ours/r_35_thermal.jpg}\\
     RGB ground truth & Thermal ground truth & Thermal ground truth & Our thermal\\
     &(Input resolution) & (High resolution) &\\
\end{tabular}
\caption{Thermal super-resolution on the \emph{hotdog} scene; the first row shows the full top-down view while the second row zooms in on the edge of the hotdog to visualize thermal super-resolution. Columns 1 \& 2 show RGB and thermal ground truth views at the training resolution, with the thermal training resolution significantly lower than the RGB training resolution to model the lower resolution of real-world thermal cameras. Column 3 shows the thermal ground truth view rendered at the same resolution as in column 1; we can render this image for our synthetic scene, but cannot measure high-resolution thermal ground truth images for our real scenes. Our thermal reconstruction, shown in column 4, is able to resolve the higher-resolution thermal details shown in column 3 by leveraging cross-channel information from the higher-resolution RGB training images.}
\label{fig:super-res}
\end{figure*}

\plotscene{sink}{00020}{Results on our \emph{sink} scene, with hot water flowing from the faucet.}

\subsection{Revealing Hidden Objects}

In \cref{fig:removal} we demonstrate an application made possible by our method, specifically by $\Loss_\density$ and the use of separate densities for visible and LWIR light. We can remove occluding objects from RGB or thermal views, thus revealing objects hidden behind other objects by rendering only the parts of the scene with RGB and thermal densities sufficiently similar to each other. Precisely, we render RGB and thermal scenes respectively with densities
\begin{align}
    \sigma_\rgb' &= \mathbbm{1}_{|\sigma_\rgb - \sigma_\therm| < \epsilon} \cdot \sigma_\rgb\\
    \sigma_\therm' &= \mathbbm{1}_{|\sigma_\rgb - \sigma_\therm| < \epsilon} \cdot \sigma_\therm
\end{align}
where $\epsilon$ is the minimum allowed magnitude of difference between the RGB and thermal densities in the rendered image. This computation is only possible due to the $\lambda_\density \Loss_\density$ term in our loss function (\cref{eq:loss}), which encourages the RGB and thermal densities to be similar.

We demonstrate this application on the \emph{pyrex} scene, which depicts a cold pack in a glass container, and the \emph{sheet} scene, which depicts a hot water kettle behind a plastic sheet. The glass container in \emph{pyrex} transmits visible but not infrared light; the plastic sheet in \emph{sheet} transmits infrared but not visible light. We show that we are able to remove the occluding material in the RGB and thermal channels respectively, revealing the shape of the object that normally would be occluded from
the ground-truth view.

\subsection{Thermal Super-resolution}

In \cref{fig:super-res}, we show qualitatively that our method achieves a level of thermal super-resolution. We demonstrate super-resolution on our synthetic \emph{hotdog} scene, for which ground truth thermal images are available at high resolution for a valid comparison. We show that with the help of the higher-resolution RGB images (column 1), despite the low resolution of the thermal training data (column 2), our reconstruction (column 4), is able to reconstruct the higher-frequency ground-truth thermal features (column 3) in the underlying volume.

\section{Conclusion}

We present a novel method that addresses critical challenges of 3D thermal reconstruction. By recovering accurate thermal camera poses through inter-camera calibration, and by integrating information from both spectra while taking into account wavelength-dependent material properties, we have achieved significant improvement in 3D thermal reconstruction both quantitatively and qualitatively.

We note that, although our implementation is built on Nerfacto, our modifications to handle thermal radiance are applicable to any radiance field model (implicit or explicit). We also note that, although we focus on the case of LWIR thermal imaging, our modifications are likely applicable to other multispectral imaging settings in which the same material interacts differently with different wavelengths. For example, our method may be relevant to multi-energy X-ray computed tomography, in which different tissues absorb different X-ray wavelengths to varying degrees, or even to RGB radiance field modeling of certain materials that absorb red, green, and blue visible wavelengths differently, such as stained glass.

% Any acknowledgments to only be included in camera ready
\ifpeerreview \else
\section*{Acknowledgments}
Many thanks to Skydio for sharing their RGB and thermal imaging of the crane structure. This material is based upon work supported by the National Science Foundation under award number 2303178 to SFK. Any opinions, findings, and conclusions or recommendations expressed in this material are those of the authors and do not necessarily reflect the views of the National Science Foundation.
\fi

\bibliographystyle{IEEEtran}
\bibliography{references}

\ifpeerreview \else
%%%% For the camera ready version, please fill out this
%%%% biography. Your camera ready should be within a 12 page limit
%%%% including acknowledgments, references and biography.

% \includegraphics command within an optional argument. (You could
% create your own custom macro containing the \includegraphics command
% to make things simpler here.)
\begin{IEEEbiography}[{\includegraphics[width=1in,height=1.25in,clip,keepaspectratio]{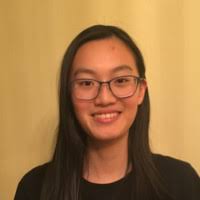}}]{Yvette Y. Lin}
is a masters student in Computer Science at Stanford University.
Her research interests lie in computational imaging, inverse problems, deep learning, and computer vision. Yvette completed her undergraduate studies at Caltech, where she majored in CS and math. 
\end{IEEEbiography}
\begin{IEEEbiography}
[{\includegraphics[width=1in,height=1.25in,clip,keepaspectratio]{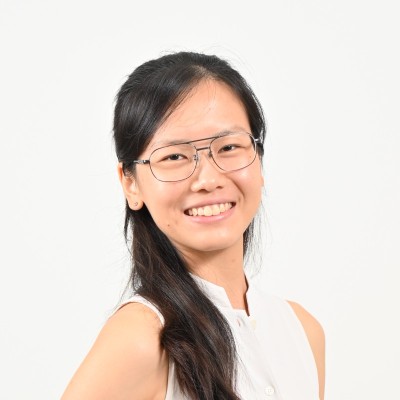}}]{Xin-Yi Pan}
is a masters student in Electrical Engineering at Stanford University, supported by the DSO Postgraduate Scholarship. Her research interests lie in the intersection of Physics and Computer Science, particularly in the areas of Optics, Quantum and Imaging. Xin-Yi completed her undergraduate studies at the National University of Singapore, where she majored in Engineering Science, specializing in Photonics and Optics. 
\end{IEEEbiography}
\begin{IEEEbiography}
[{\includegraphics[width=1in,height=1.25in,clip,keepaspectratio]{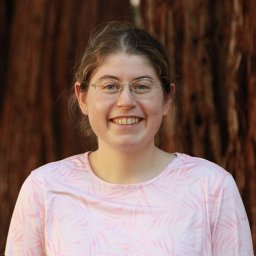}}]{Sara Fridovich-Keil}
is a postdoctoral fellow at Stanford University, where she works on foundations and applications of machine learning and signal processing in computational imaging. She is currently supported by an NSF Mathematical Sciences Postdoctoral Research Fellowship. Sara received her PhD in Electrical Engineering and Computer Sciences in 2023 from UC Berkeley, where she was supported by an NSF Graduate Research Fellowship. Sara is an incoming Assistant Professor in Electrical and Computer Engineering at Georgia Tech.
\end{IEEEbiography}
\begin{IEEEbiography}
[{\includegraphics[width=1in,height=1.25in,clip,keepaspectratio]{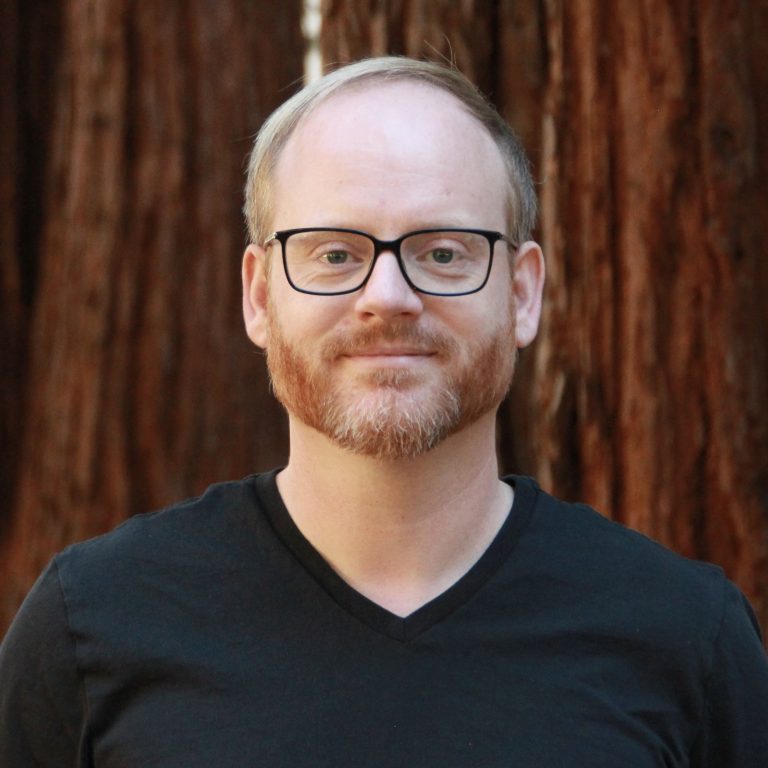}}]{Gordon Wetzstein}
is an Associate Professor of Electrical Engineering and, by courtesy, of Computer Science at Stanford University. He is the leader of the Stanford Computational Imaging Lab and a faculty co-director of the Stanford Center for Image Systems Engineering. At the intersection of computer graphics and vision, artificial intelligence, computational optics, and applied vision science, Prof. Wetzstein's research has a wide range of applications in next-generation imaging, wearable computing, and neural rendering systems. Prof. Wetzstein is a Fellow of Optica and the recipient of numerous awards, including an IEEE VGTC Virtual Reality Technical Achievement Award, an NSF CAREER Award, an Alfred P. Sloan Fellowship, an ACM SIGGRAPH Significant New Researcher Award, a Presidential Early Career Award for Scientists and Engineers (PECASE), an SPIE Early Career Achievement Award, an Electronic Imaging Scientist of the Year Award, an Alain Fournier Ph.D. Dissertation Award as well as many Best Paper and Demo Awards.
\end{IEEEbiography}

%\newpage


%\vfill
\clearpage
\fi

\fi

\ifsupplement
\clearpage
\section*{Supplement}
\setcounter{section}{0}

\ifmain
\else
\setcounter{figure}{13}
\setcounter{table}{1}
\fi

\section{Detailed results}

In \cref{tab:detailed-results} we present the quantitative results for each individual scene, as well as quantitative ablations excluding each of our proposed regularizers. \Cref{fig:ablations} visualizes the value of these regularizers via qualitative ablations.

\begin{figure}[h]
    \centering
    \begin{tabular}{c@{}c@{}c}
        Ground Truth & Ours & w/o $\Loss_\crosschannel$\\
        \drawbox{\includegraphics[width=0.32\linewidth]{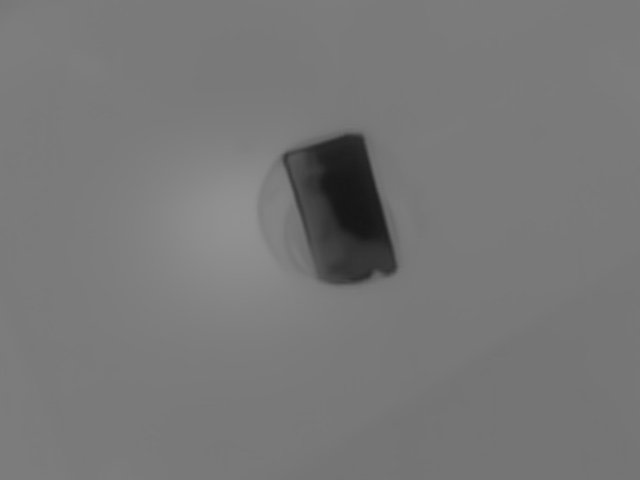}}{0.4}{0.02}{0.95}{0.23}&
        \drawbox{\includegraphics[width=0.32\linewidth]{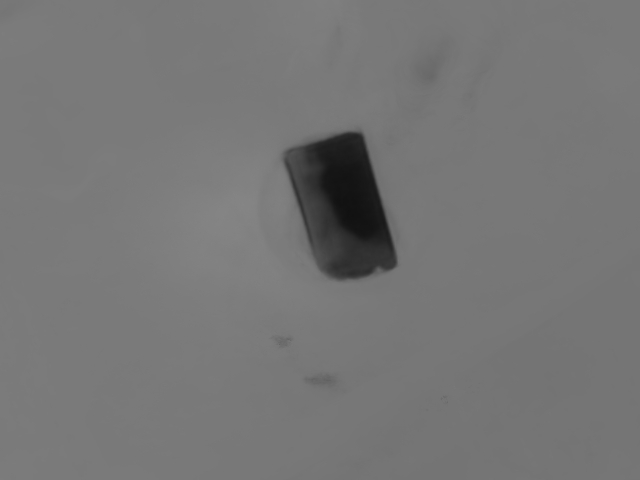}}{0.4}{0.02}{0.95}{0.23}&
        \drawbox{\includegraphics[width=0.32\linewidth]{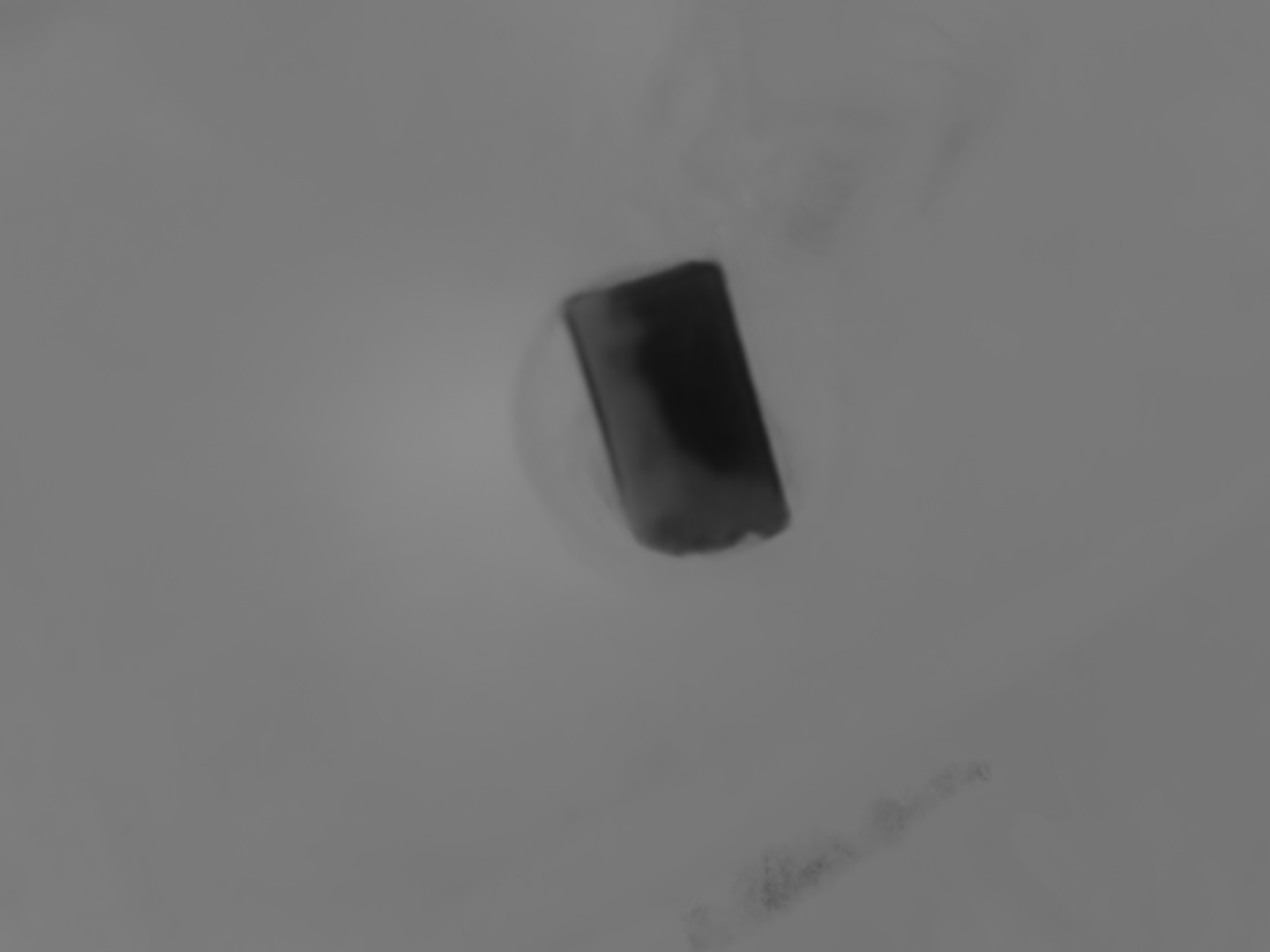}}{0.4}{0.02}{0.95}{0.23}\\
        Ground Truth & Ours & w/o $\Loss_\density$\\
        \includegraphics[width=0.32\linewidth]{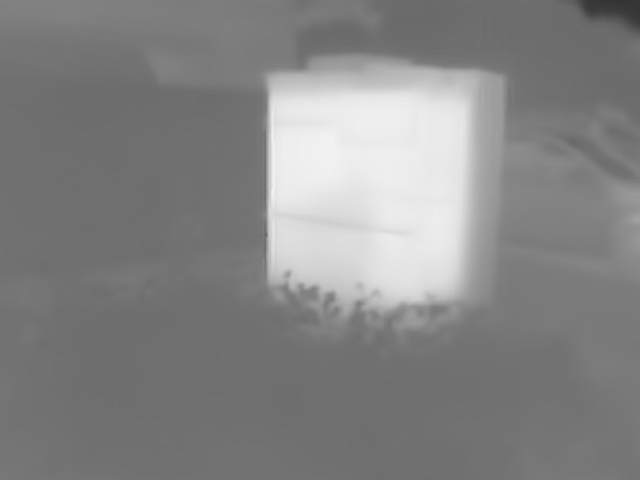}&
        \includegraphics[width=0.32\linewidth]{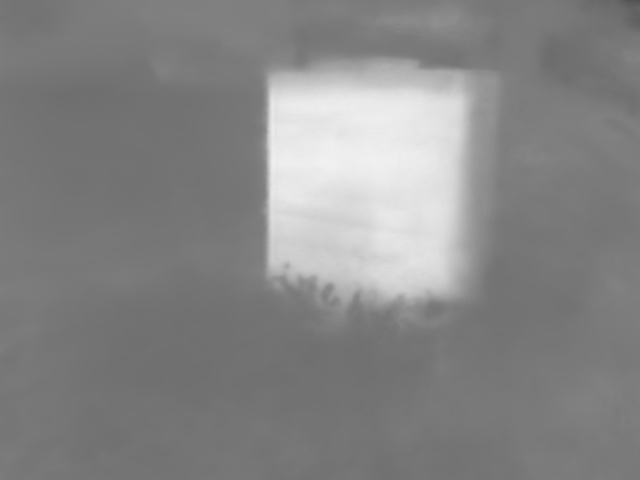}&
        \includegraphics[width=0.32\linewidth]{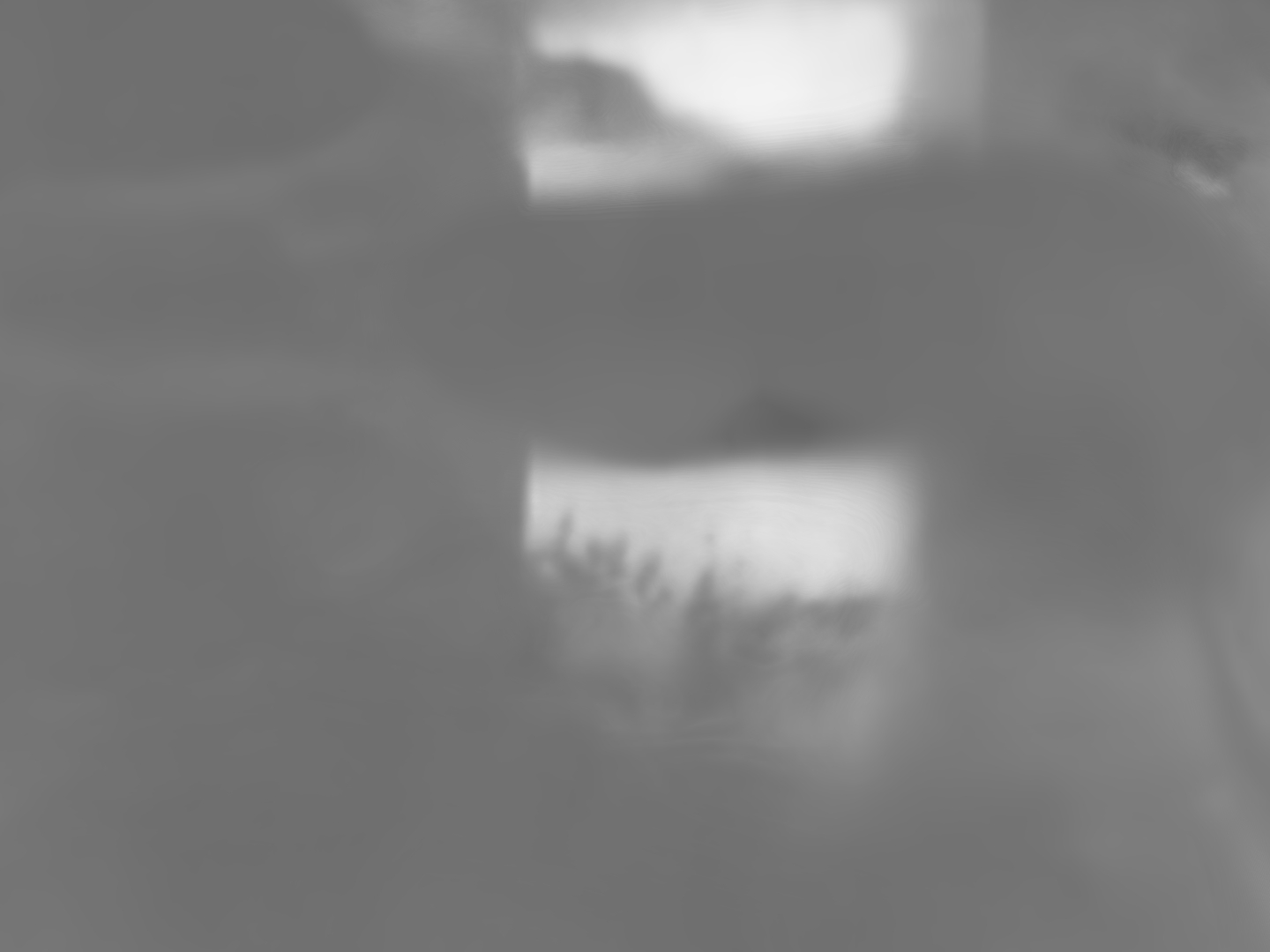}\\
        Ground Truth & Ours & w/o $\Loss_\tv$\\
        \drawbox{\adjincludegraphics[width=0.32\linewidth]{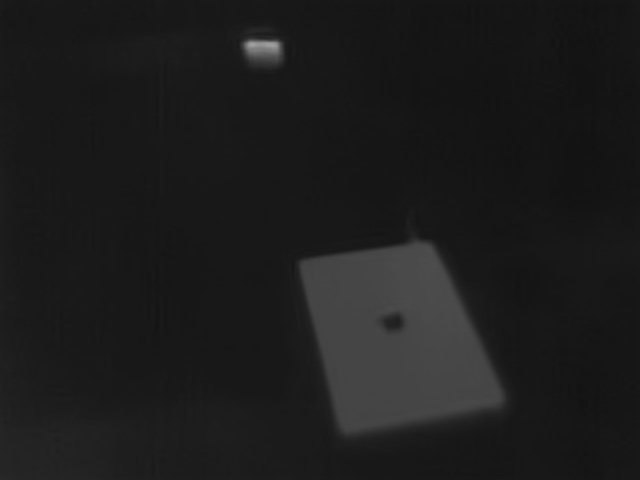}}{0.3}{0.78}{0.5}{0.98}&
        \drawbox{\adjincludegraphics[width=0.32\linewidth]{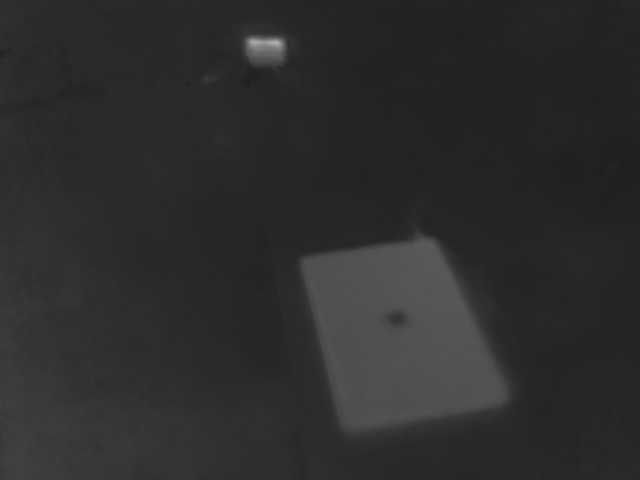}}{0.3}{0.78}{0.5}{0.98}&
        \drawbox{\adjincludegraphics[width=0.32\linewidth]{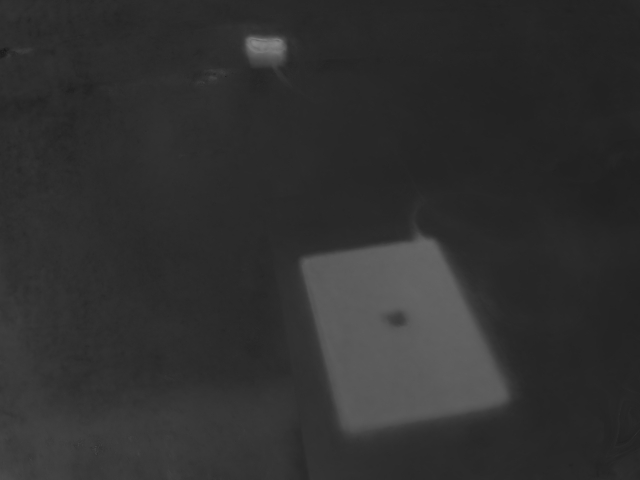}}{0.3}{0.78}{0.5}{0.98}\\
        \addlinespace[-0.8ex]
        \adjincludegraphics[trim={{0.3\width} {0.8\height} {0.5\width} {0\height}}, clip, width=0.32\linewidth]{figs/charger/ground_truth_thermal/frame_00034.jpg}&
        \adjincludegraphics[trim={{0.3\width} {0.8\height} {0.5\width} {0\height}}, clip, width=0.32\linewidth]{figs/charger/ours/frame_00034_thermal.jpg}&
        \adjincludegraphics[trim={{0.3\width} {0.8\height} {0.5\width} {0\height}}, clip, width=0.32\linewidth]{figs/charger/no-pixelwise-tv_thermal/frame_00034.jpg}
    \end{tabular}
    \caption{Qualitative ablations showing benefit from each of our regularizers.
    Row 1: A thermal test view from \textit{pyrex}; note the appearance of artifacts without the cross-channel loss.
    Row 2: A thermal test view from \textit{generator}; without the density regularization, the geometry of the thermal scene is incorrect, including an incorrect generator shape and occlusions.
    Rows 3 \& 4: A thermal test view from \textit{charger}; w/o the total variation loss, some hallucinated high-frequency features, such as ring-like patterns on the charger, appear. Additionally, though this is more obvious by observing the rendered result at high resolution, the full image w/o $\Loss_\tv$ exhibits a rough texture not found in the ground-truth image.
   }
    \label{fig:ablations}
\end{figure}

\begin{table*}[!h]
\renewcommand{\arraystretch}{1.3}
\caption{Per-scene results.}
\begin{center}
\begin{tabular}{lccccccccccc}
\hline
Method &&& \emph{pyrex} & \emph{heater} & \emph{sink} & \emph{charger} & \emph{trace} & \emph{generator} & \emph{generators} & \emph{sheet} & \emph{engine}\\
\hline
Ours & PSNR($\uparrow$) & RGB & 22.12 & 24.88 & 28.31 & 25.09 & 22.59 & 20.44 & 18.84 & 24.84 & 22.25\\
&& Thermal & 35.78 & 35.24 & 23.83 & 32.60 & 30.92 & 32.80 & 31.13 & 27.36 & 31.98\\
& SSIM($\uparrow$) & RGB & 0.715 & 0.827 & 0.911 & 0.890 & 0.851 & 0.610 & 0.436 & 0.800 & 0.804\\
&& Thermal & 0.987 & 0.980 & 0.890 & 0.956 & 0.960 & 0.977 & 0.914 & 0.914 & 0.925\\
& LPIPS($\downarrow$) & RGB & 0.439 & 0.370 & 0.274 & 0.321 & 0.385 & 0.504 & 0.571 & 0.360 & 0.259\\
&& Thermal & 0.022 & 0.018 & 0.120 & 0.044 & 0.044 & 0.047 & 0.066 & 0.060 & 0.076 \\
\hline
Nerfacto \{RGB\}\{T\} & PSNR($\uparrow$) & RGB & 21.30 & 24.31 & 27.26 & 23.65 & 22.14 & 19.24 & 18.15 & 24.33 & 20.81\\
&& Thermal & 22.04 & 19.60 & 13.40 & 16.80 & 23.89 & 16.43 & 21.70 & 19.46 & 19.81\\
& SSIM($\uparrow$) & RGB & 0.689 & 0.805 & 0.894 & 0.853 & 0.838 & 0.533 & 0.389 & 0.780 & 0.751\\
&& Thermal & 0.970 & 0.781 & 0.796 & 0.757 & 0.948 & 0.951 & 0.958 & 0.883 & 0.793\\
& LPIPS($\downarrow$) & RGB & 0.449 & 0.370 & 0.278 & 0.338 & 0.378 & 0.533 & 0.597 & 0.365 & 0.268\\
&& Thermal & 0.096 & 0.179 & 0.476 & 0.271 & 0.119 & 0.133 & 0.153 & 0.185 & 0.425 \\
\hline
Nerfacto \{RGBT\} & PSNR($\uparrow$) & RGB & 21.76 & 23.68 & 23.89 & 24.56 & 22.90 & 19.02 & 18.18 & 22.89 & 20.19\\
&& Thermal & 32.31 & 20.94 & 15.65 & 13.95 & 22.04 & 29.87 & 17.33 & 19.71 & 21.77\\
& SSIM($\uparrow$) & RGB & 0.713 & 0.806 & 0.845 & 0.879 & 0.843 & 0.529 & 0.394 & 0.748 & 0.728\\
&& Thermal & 0.985 & 0.668 & 0.754 & 0.655 & 0.949 & 0.982 & 0.936 & 0.812 & 0.864\\
& LPIPS($\downarrow$) & RGB & 0.457 & 0.380 & 0.341 & 0.327 & 0.388 & 0.570 & 0.604 & 0.390 & 0.293\\
&& Thermal & 0.050 & 0.182 & 0.276 & 0.349 & 0.053 & 0.068 & 0.231 & 0.238 & 0.188 \\
\hline
Ours /wo $\Loss_\crosschannel$ & PSNR($\uparrow$) & RGB & 22.28 & 24.91 & 28.20 & 25.08 & 22.64 & 20.28 & 18.69 & 24.92 & 22.21\\
&& Thermal & 34.97 & 35.20 & 21.99 & 27.03 & 35.45 & 30.88 & 32.34 & 24.31 & 32.61\\
& SSIM($\uparrow$) & RGB & 0.721 & 0.829 & 0.910 & 0.900 & 0.851 & 0.605 & 0.425 & 0.798 & 0.801\\
&& Thermal & 0.987 & 0.984 & 0.893 & 0.921 & 0.960 & 0.974 & 0.920 & 0.791 & 0.925\\
& LPIPS($\downarrow$) & RGB & 0.437 & 0.368 & 0.271 & 0.323 & 0.380 & 0.507 & 0.570 & 0.360 & 0.259\\
&& Thermal & 0.023 & 0.015 & 0.143 & 0.083 & 0.042 & 0.057 & 0.059 & 0.080 & 0.076\\
\hline
Ours w/o $\Loss_\density$ & PSNR($\uparrow$) & RGB & 22.16 & 24.97 & 28.51 & 25.10 & 22.76 & 20.45 & 18.73 & 24.89 & 21.93\\
&& Thermal & 34.81 & 33.64 & 20.08 & 31.07 & 30.37 & 28.61 & 30.24 & 28.83 & 30.24\\
& SSIM($\uparrow$) & RGB & 0.715 & 0.828 & 0.912 & 0.887 & 0.852 & 0.607 & 0.429 & 0.795 & 0.794\\
&& Thermal & 0.987 & 0.974 & 0.722 & 0.947 & 0.958 & 0.958 & 0.915 & 0.961 & 0.934\\
& LPIPS($\downarrow$) & RGB & 0.439 & 0.360 & 0.269 & 0.313 & 0.381 & 0.502 & 0.562 & 0.354 & 0.260\\
&& Thermal & 0.024 & 0.020 & 0.153 & 0.051 & 0.044 & 0.087 & 0.063 & 0.042 & 0.078 \\
\hline
Ours w/o $\Loss_\tv$ & PSNR($\uparrow$) & RGB & 22.35 & 24.97 & 28.58 & 25.25 & 22.61 & 20.36 & 18.72 & 24.82 & 22.29\\
&& Thermal & 36.01 & 35.53 & 25.66 & 31.18 & 30.05 & 32.90 & 29.52 & 23.97 & 29.89\\
& SSIM($\uparrow$) & RGB & 0.717 & 0.831 & 0.912 & 0.892 & 0.852 & 0.602 & 0.428 & 0.796 & 0.803\\
&& Thermal & 0.987 & 0.984 & 0.894 & 0.950 & 0.957 & 0.976 & 0.912 & 0.786 & 0.932\\
& LPIPS($\downarrow$) & RGB & 0.443 & 0.370 & 0.271 & 0.319 & 0.383 & 0.501 & 0.567 & 0.359 & 0.259\\
&& Thermal & 0.024 & 0.015 & 0.115 & 0.057 & 0.046 & 0.047 & 0.069 & 0.089 & 0.073 \\
\hline
\end{tabular}
\end{center}
\label{tab:detailed-results}
\end{table*}

\plotscene{generators}{00021}{Results on our \emph{generators} scene.}

\section{Additional qualitative results}

In \cref{fig:generators} we present qualitative results on our \emph{generators} scene.

In \cref{fig:multiview_pyrex} we present RGB and thermal renderings from multiple viewpoints of our \emph{pyrex} scene, to demonstrate multiview consistency. Video results are also available on the project webpage.

\begin{figure*}[!h]
\centering
\begin{tabular}{l@{~~}c@{}c@{}c@{}c}
    \rotatebox{90}{~~~~~~~~~~GT RGB} &
    \includegraphics[width=0.23\linewidth]{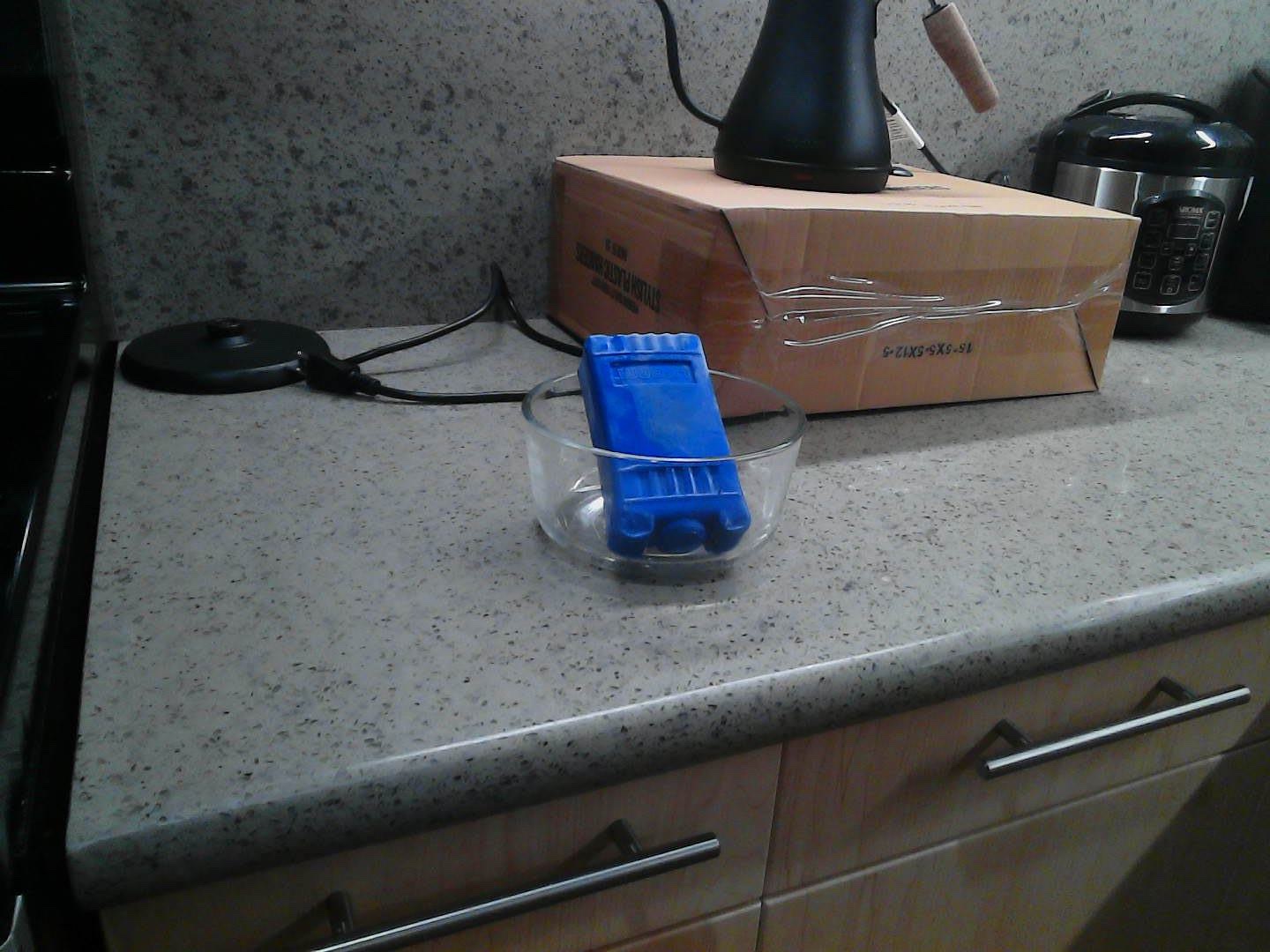}&
    \includegraphics[width=0.23\linewidth]{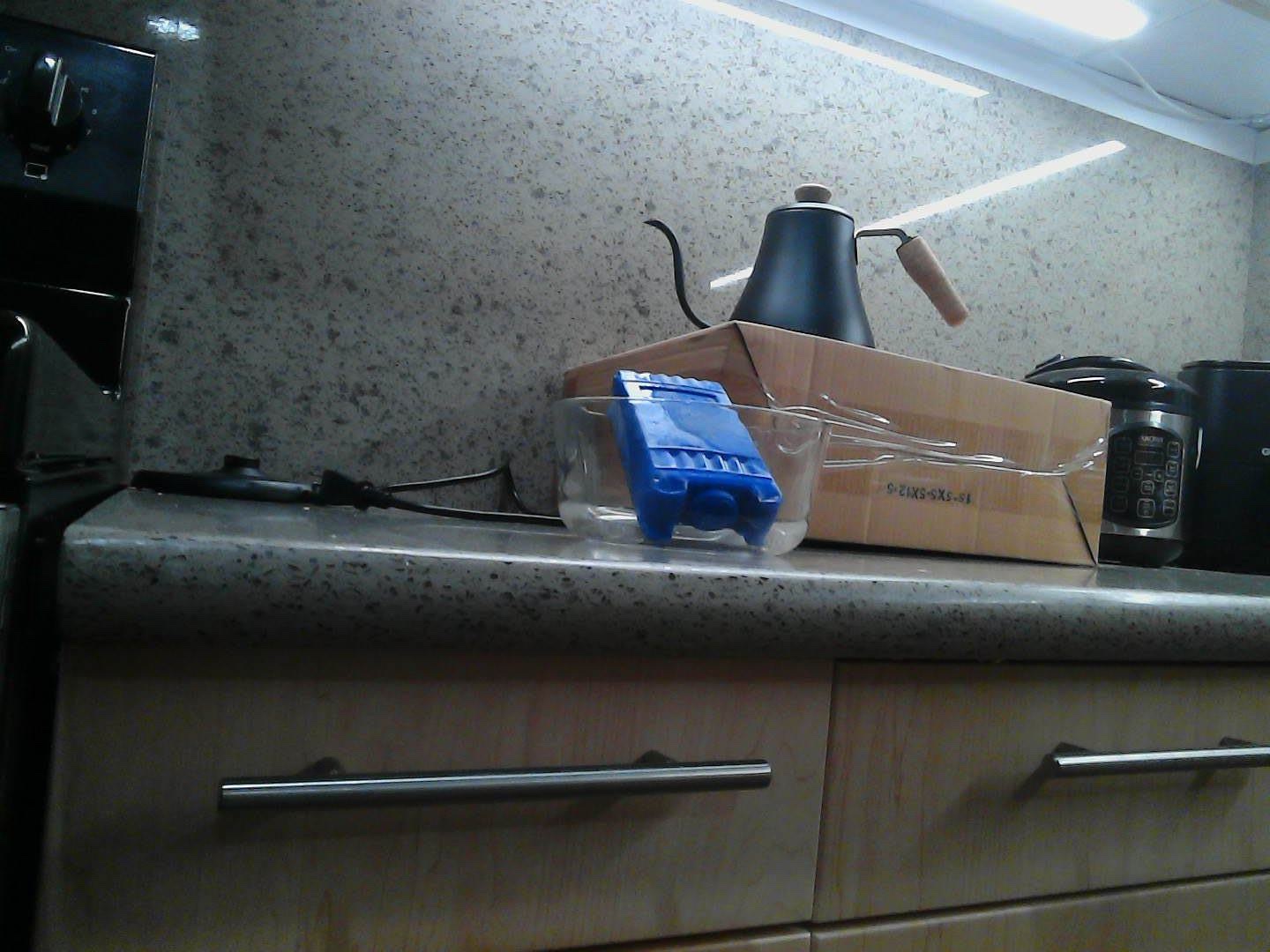}&
    \includegraphics[width=0.23\linewidth]{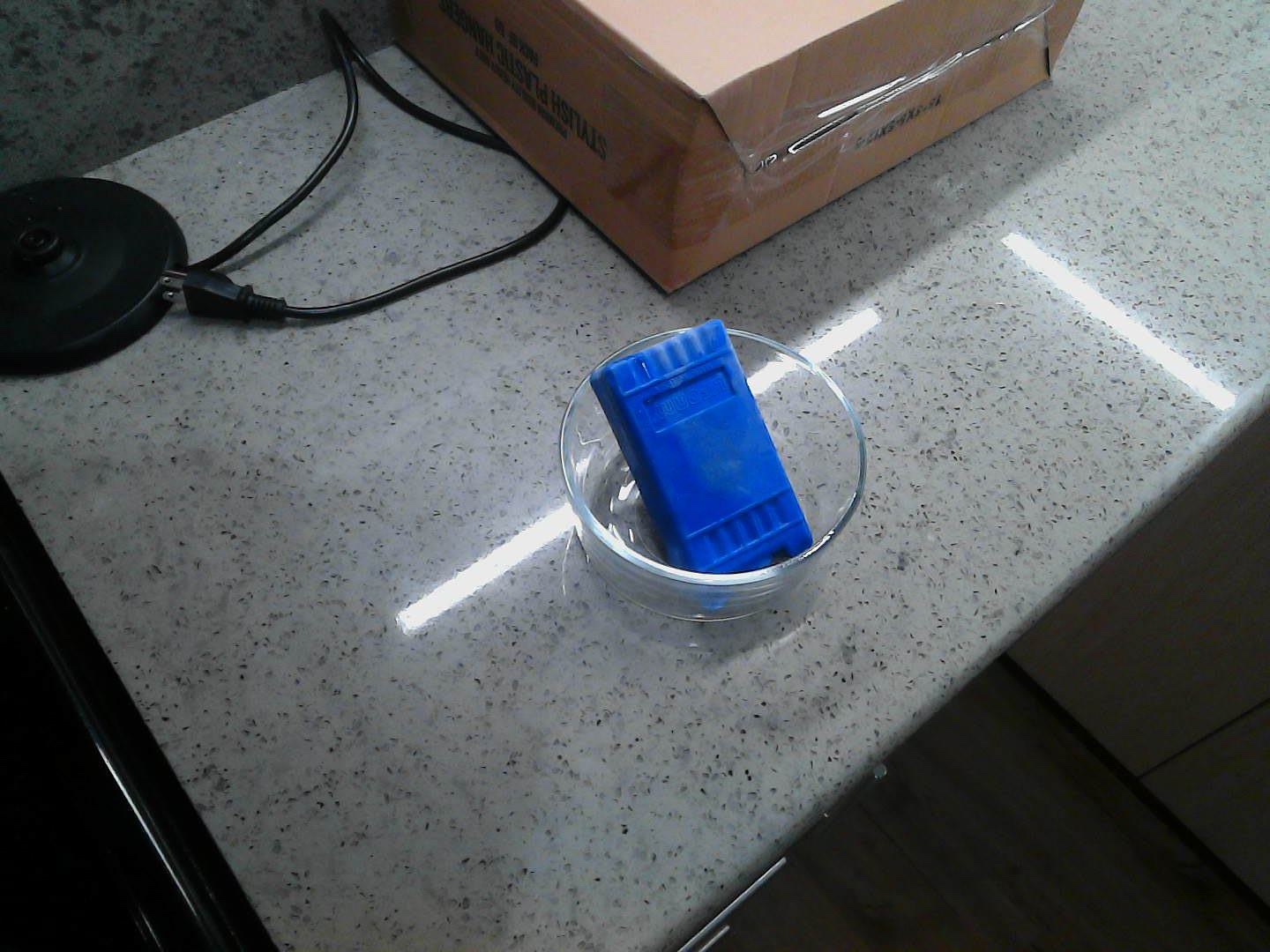}&
    \includegraphics[width=0.23\linewidth]{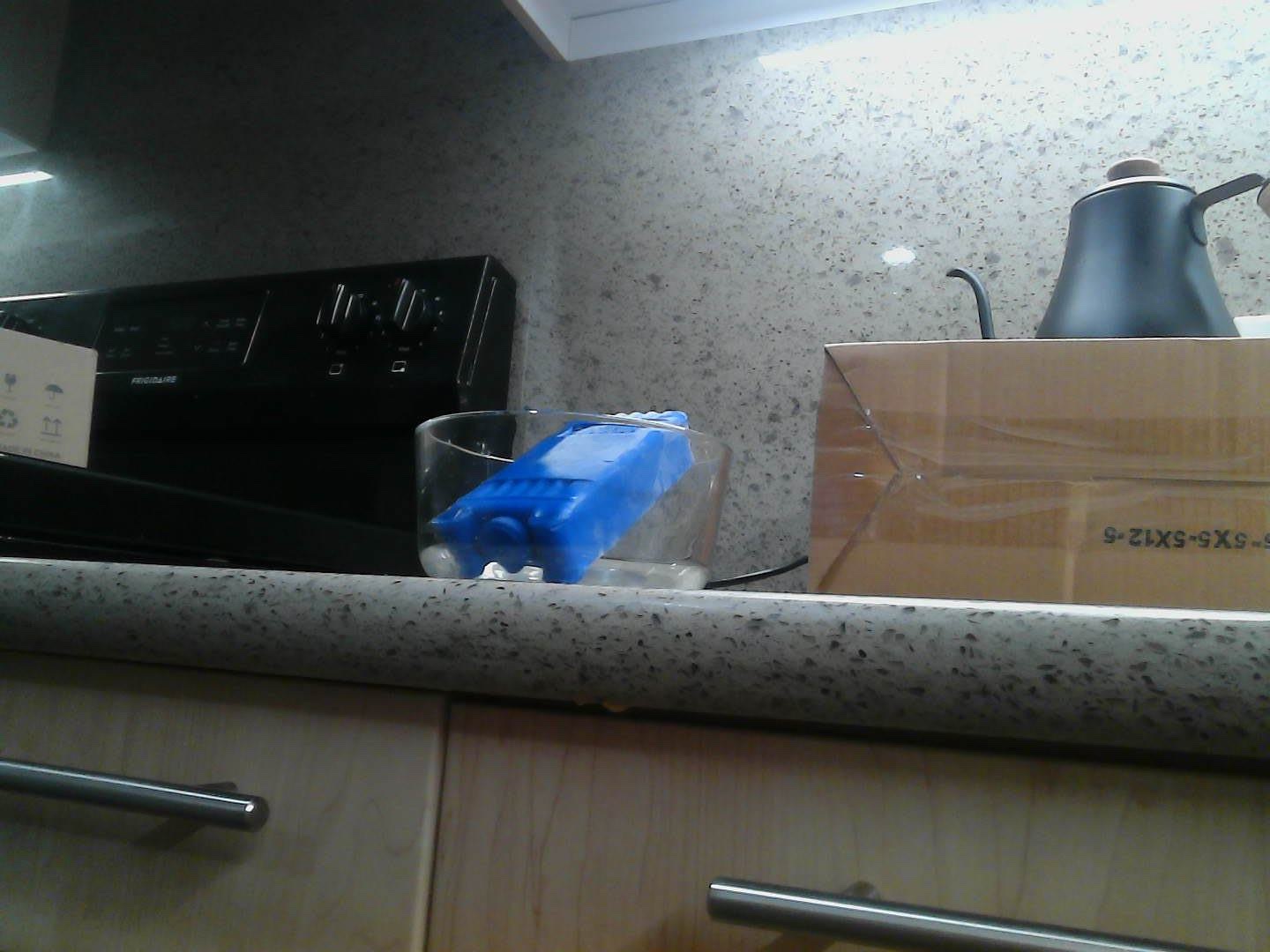}\\
    \addlinespace[-0.8ex]
    \rotatebox{90}{~~~~~~~~~~Our RGB} &
    \includegraphics[width=0.23\linewidth]{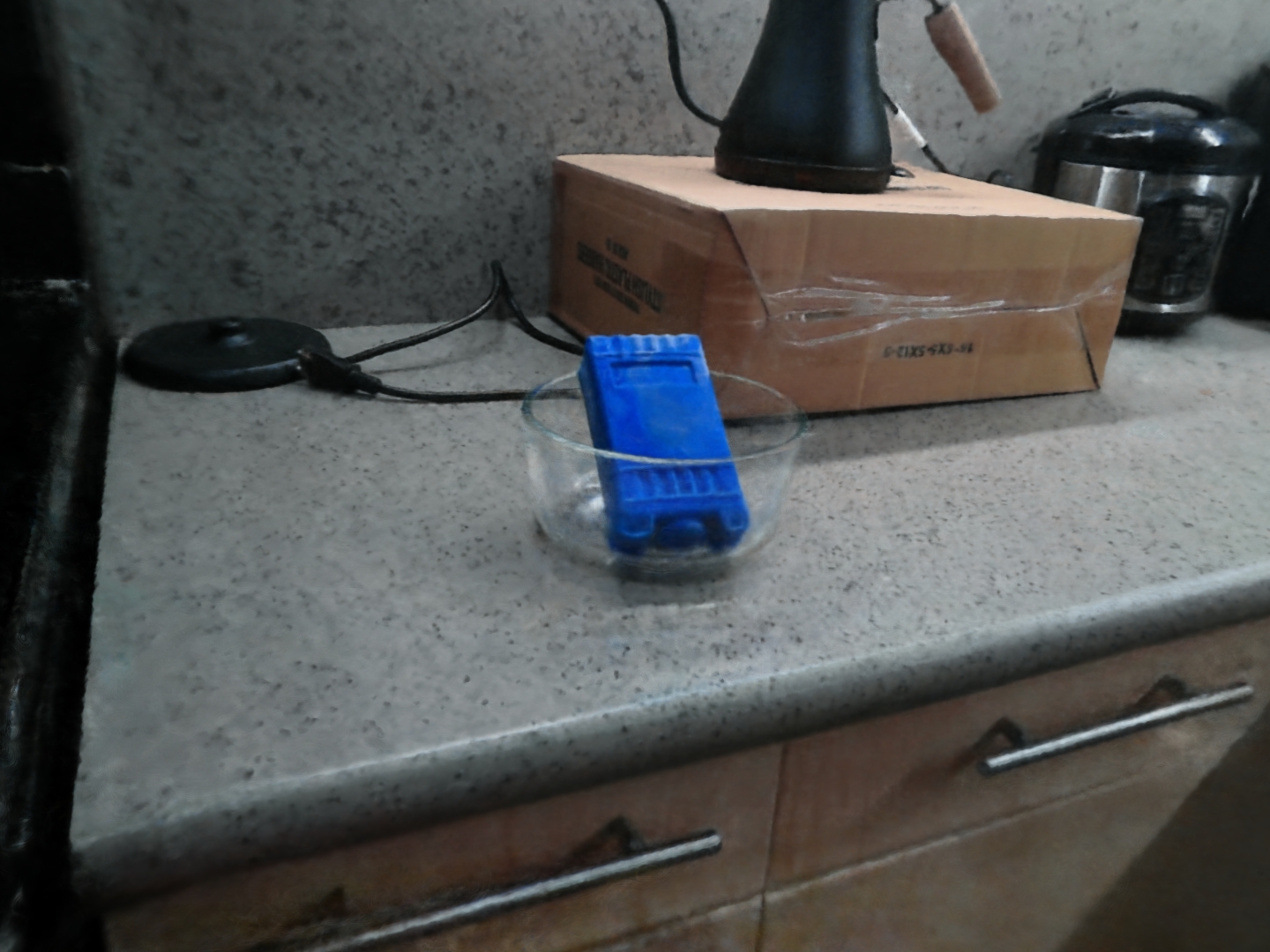}&
    \includegraphics[width=0.23\linewidth]{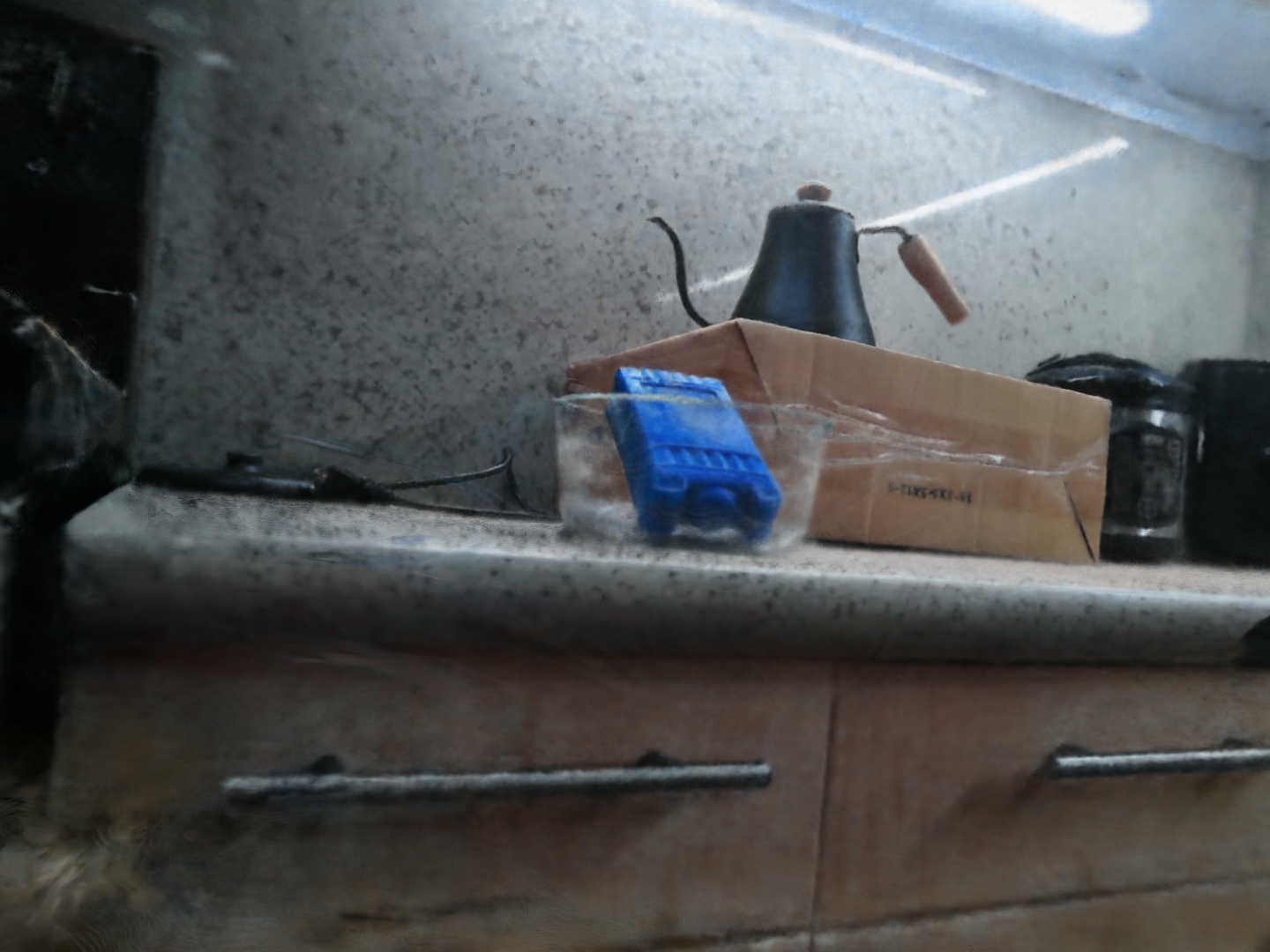}&
    \includegraphics[width=0.23\linewidth]{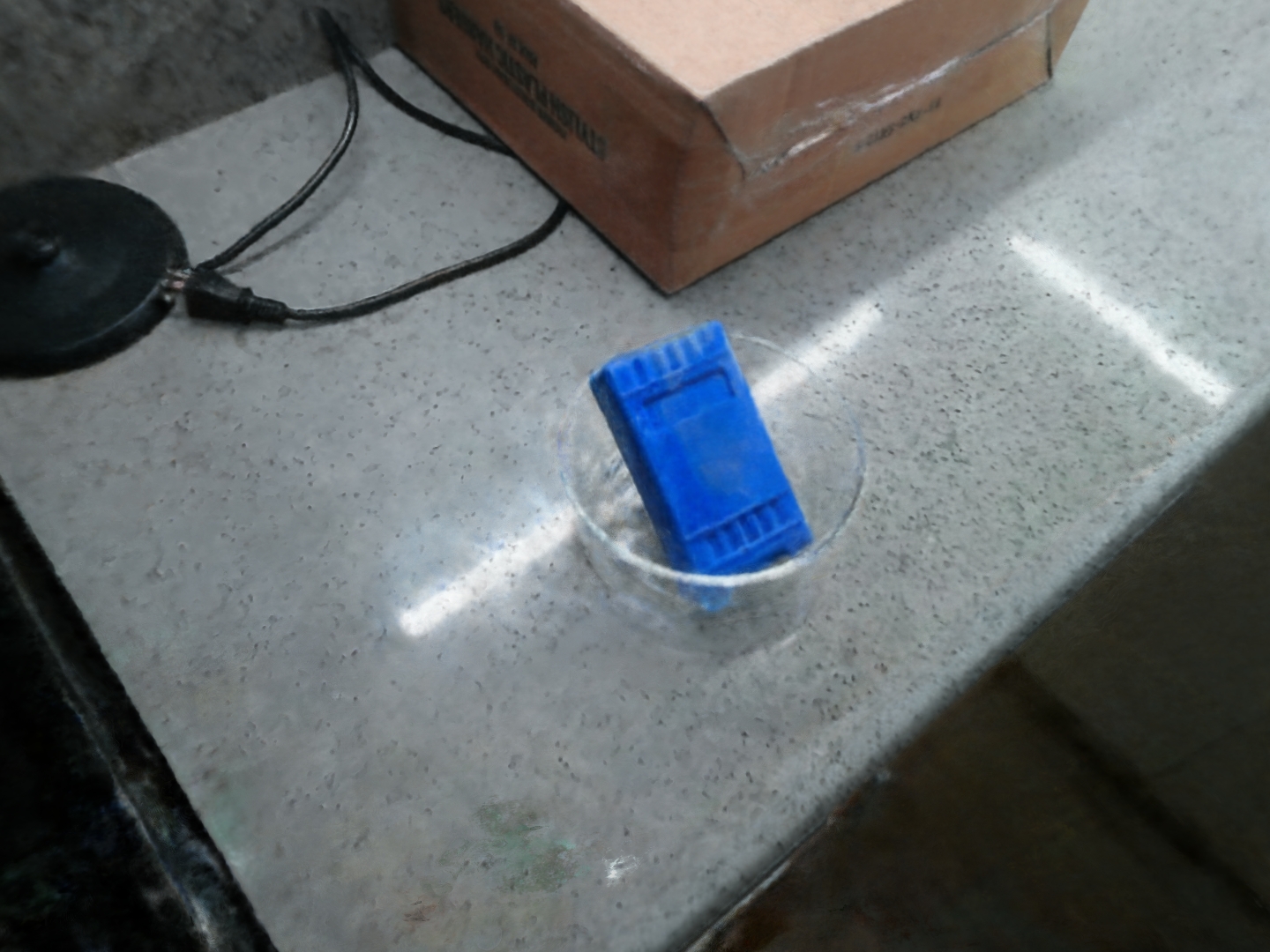}&
    \includegraphics[width=0.23\linewidth]{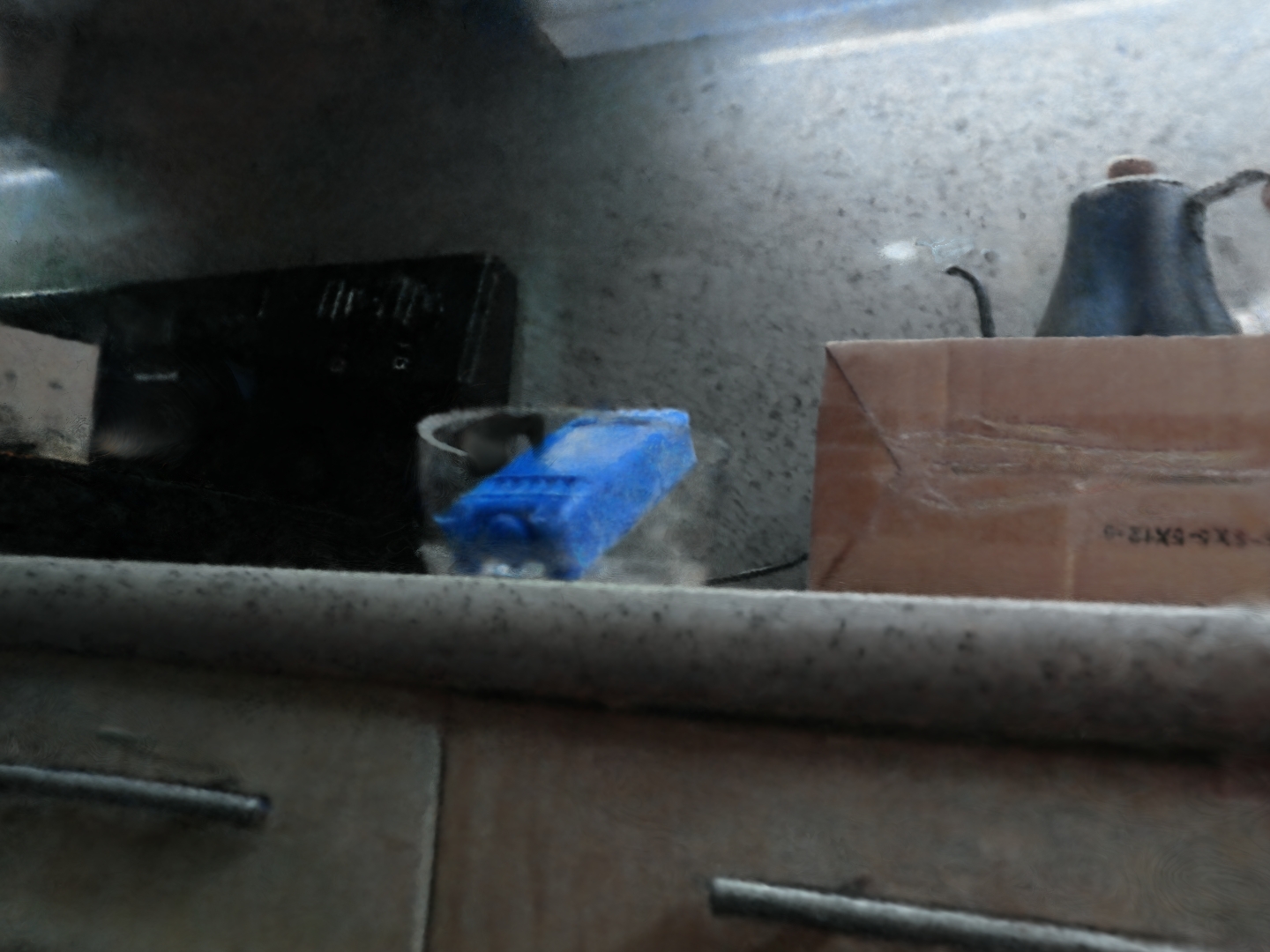}\\
    \rotatebox{90}{~~~~~~GT Thermal} &
    \includegraphics[width=0.23\linewidth]{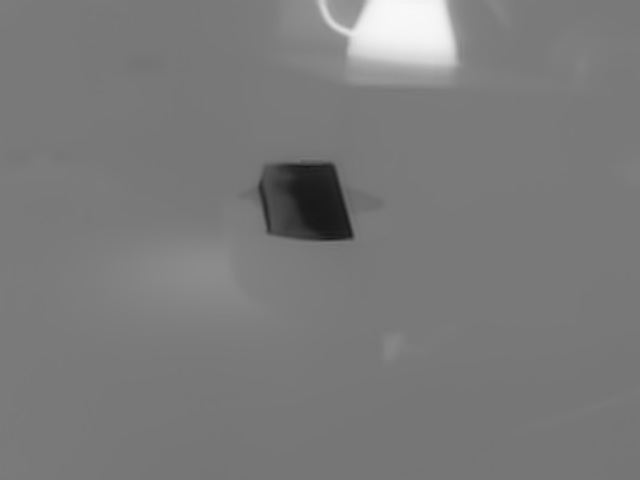}&
    \includegraphics[width=0.23\linewidth]{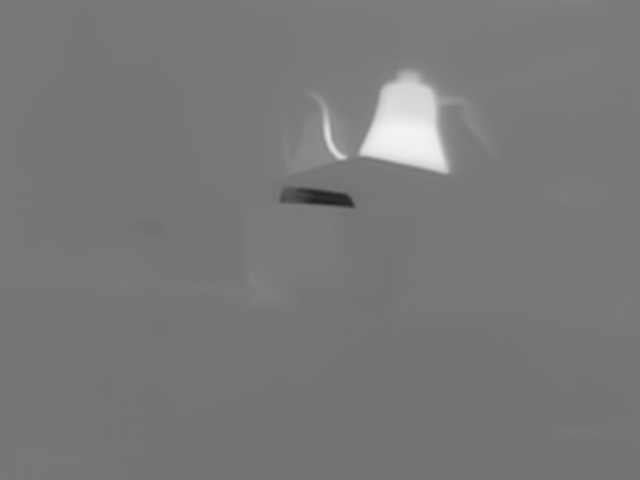}&
    \includegraphics[width=0.23\linewidth]{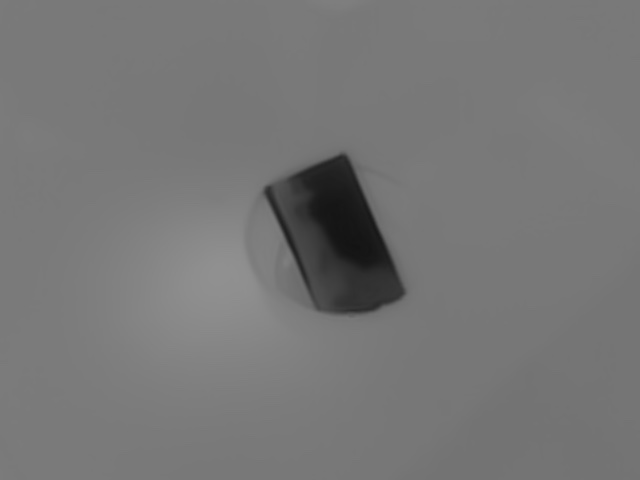}&
    \includegraphics[width=0.23\linewidth]{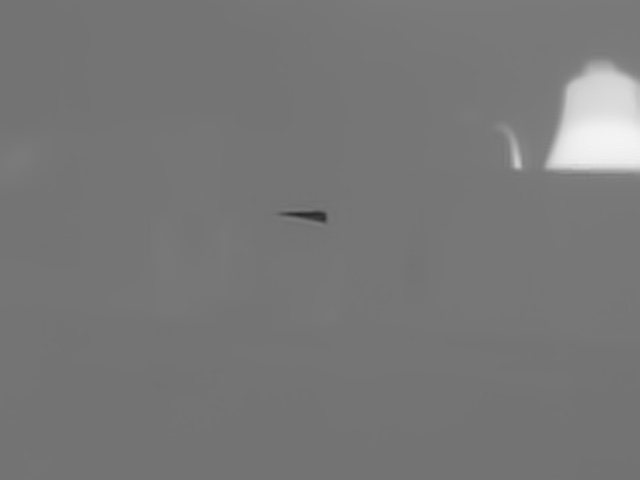}\\
    \addlinespace[-0.8ex]
    \rotatebox{90}{~~~~~~Our Thermal} &
    \includegraphics[width=0.23\linewidth]{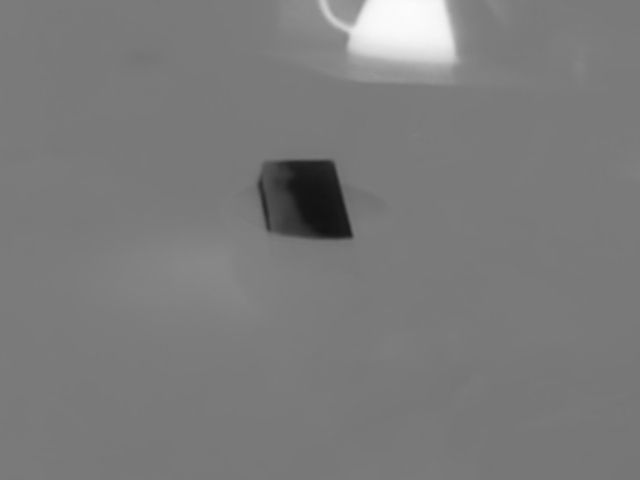}&
    \includegraphics[width=0.23\linewidth]{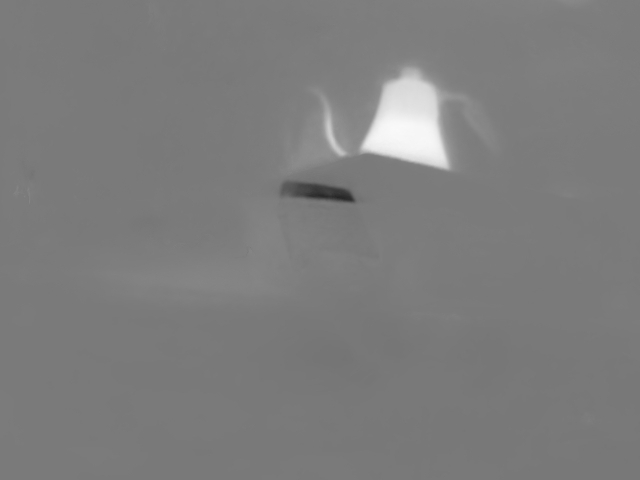}&
    \includegraphics[width=0.23\linewidth]{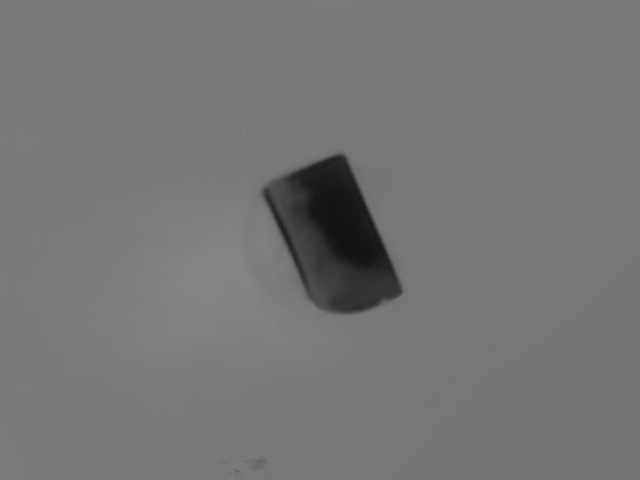}&
    \includegraphics[width=0.23\linewidth]{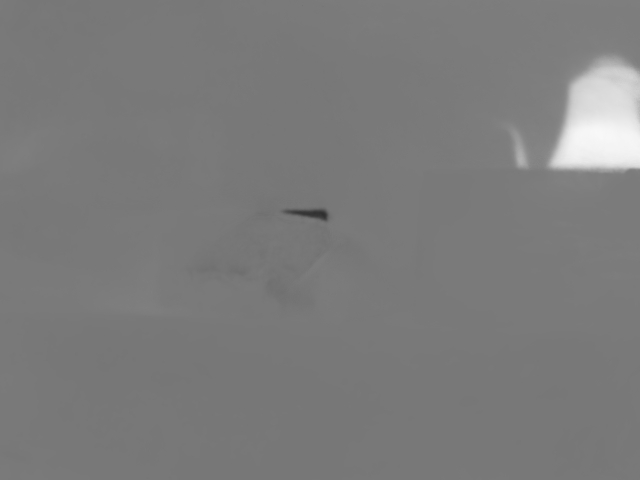}
\end{tabular}
\caption{Multiple views of our reconstructed \emph{pyrex} scene, demonstrating the multiview consistency of our method. 
}
\label{fig:multiview_pyrex}
\end{figure*}

\fi

\end{document}